\documentclass[letterpaper]{article} 
\usepackage{aaai25} 
\usepackage{times}  
\usepackage{helvet}  
\usepackage{courier}  
\usepackage[hyphens]{url}  
\usepackage{graphicx} 
\usepackage{natbib}  
\usepackage{caption} 
\usepackage{algorithm}
\usepackage{algorithmic}

\usepackage{amssymb} 
\usepackage{amsmath}
\usepackage{bm}
\usepackage{booktabs}
\usepackage{pifont}
\usepackage{tikz}
\definecolor{skyblue}{rgb}{0.53, 0.81, 0.92}
\definecolor{forestgreen}{rgb}{0.13, 0.55, 0.13}
\definecolor{springgreen}{rgb}{0.0, 1.0, 0.5}
\usepackage{threeparttable}
\usepackage{xcolor}
\definecolor{purplecolor}{HTML}{5A58CD}
\definecolor{redcolor}{HTML}{FF605C}

\definecolor{6_1}{HTML}{C0DFAA}
\definecolor{6_2}{HTML}{CDCBFE}
\definecolor{6_3}{HTML}{F5CFA5}
\definecolor{6_4}{HTML}{E59F99}

\usepackage{newfloat}
\usepackage{listings}
\DeclareCaptionStyle{ruled}{labelfont=normalfont,labelsep=colon,strut=off} 
\floatstyle{ruled}
\newfloat{listing}{tb}{lst}{}
\floatname{listing}{Listing}
\title{GNN-SKAN: Harnessing the Power of SwallowKAN to Advance Molecular Representation Learning with GNNs}

{\author{
Ruifeng Li\textsuperscript{\rm 1,2},
Mingqian Li\textsuperscript{\rm 2},
Wei Liu\textsuperscript{\rm 3},
Hongyang Chen\textsuperscript{\rm 2*}
}}
\affiliations{
    \textsuperscript{\rm 1}College of Computer Science and Technology, Zhejiang University\\

    \textsuperscript{\rm 2}Zhejiang Lab\\
    \textsuperscript{\rm 3}Shanghai Artificial Intelligence Laboratory\\
    lirf@zju.edu.cn, mingqian.li@zhejianglab.com,
    captain.130@sjtu.edu.cn,
    dr.h.chen@ieee.org
}

\usepackage{bibentry}

\begin{document}

\maketitle

\begin{abstract}
Effective molecular representation learning is crucial for advancing molecular property prediction and drug design. 
Mainstream molecular representation learning approaches are based on Graph Neural Networks (GNNs).
However, these approaches struggle with three significant challenges: insufficient annotations, molecular diversity, and architectural limitations such as over-squashing, which leads to the loss of critical structural details. 
%
%
%
%
%
%
To address these challenges, we introduce a new class of GNNs that integrates the Kolmogorov-Arnold Networks (KANs), known for their robust data-fitting
capabilities and high accuracy in small-scale AI + Science
tasks. By incorporating KANs into GNNs, our model enhances the representation of molecular structures. 
We further advance this approach with a variant called \textbf{\underline{S}wallow\underline{KAN}} (\textbf{SKAN}), which employs adaptive Radial Basis Functions (RBFs) as the core of the non-linear neurons.
This innovation improves both computational efficiency and adaptability to diverse molecular structures.
Building on the strengths of SKAN, we propose a new class of GNNs, \textbf{GNN-SKAN}, and its augmented variant, \textbf{GNN-SKAN+}, which incorporates a SKAN-based classifier to further boost performance.
To our knowledge, this is the first work to integrate KANs into GNN architectures tailored for molecular representation learning. 
Experiments across 6 classification datasets, 6 regression datasets, and 4 few-shot learning datasets demonstrate that our approach achieves new state-of-the-art performance in terms of accuracy and computational cost. The code is available at \url{https://github.com/Lirain21/GNN-SKAN.git}.

%
%

%


\end{abstract}

\begin{figure}
    \centering
    \includegraphics[width=1.0\linewidth]{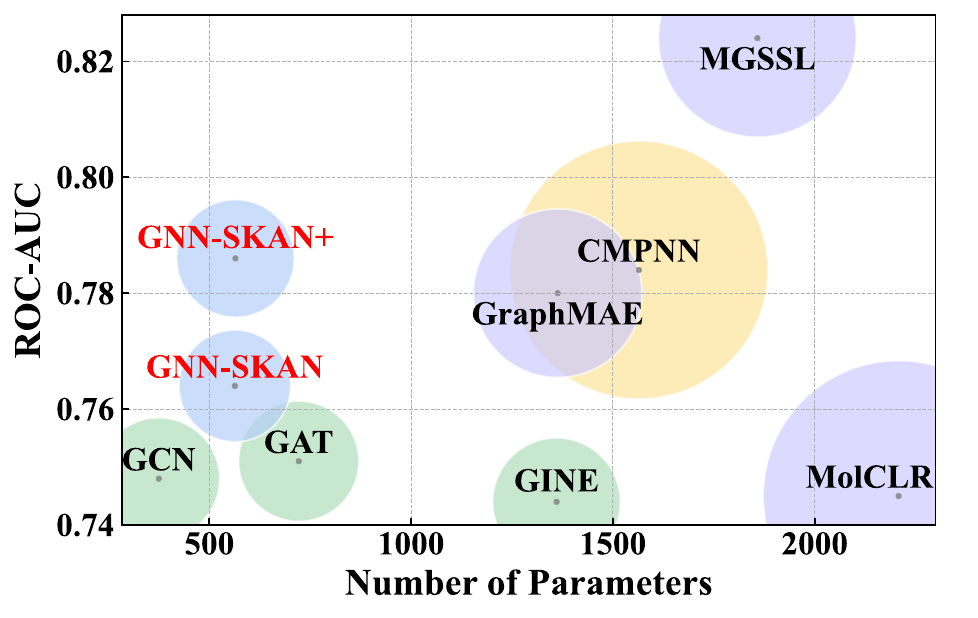}
    \caption{The visualization of our models' performance compared to the SOTAs on the HIV dataset. The circle size represents per-epoch computational cost. Purple circles represent self-supervised methods, green and yellow circles represent supervised methods, and blue circles represent our proposed methods (GNN-SKAN, GNN-SKAN+).}
    \label{fig:hiv_visualization}
\end{figure}

\section{Introduction}
Precisely predicting the physicochemical properties of molecules \cite{livingstone2000characterization, li2022deep}, drug activities \cite{datta2004crystal, sadybekov2023computational}, and material characteristics \cite{pilania2021machine, jablonka2024leveraging} is essential for advancing drug discovery and material science. The success of these predictions significantly hinges on the quality of molecular representation learning, which transforms molecular structures into high-dimensional representations \cite{fang2022geometry, yang2019analyzing}. 
However, there are two significant challenges in this field: 1) a lack of adequate labeled data due to low screening success rates \cite{wieder2020compact}; 2) wide variations in molecular structures, compositions, and properties, which reflect molecular diversity \cite{guo2022polygrammar}.

In molecular representation learning, mainstream GNN-based methods \cite{yi2022graph, fang2022geometry, yu2022molecular, chen2024uncovering, pei2024hago} model the topological structure of molecules by representing atoms as nodes and bonds as edges. 
Most approaches commonly employ the message-passing (MP) mechanism \cite{gilmer2020message}, known as MP-GNNs, to  aggregate and update representations across molecular graphs. 
Notable efforts include Graph Convolutional Network (GCN) \cite{kipf2017semisupervised}, Graph Attention Network (GAT) \cite{velivckovic2018graph}, and 
 Graph Isomorphism Network (GIN) \cite{xu2018powerful}. 
Remarkably, interactions among distant atoms or functional groups within a molecule are crucial for molecular representation learning \cite{xiong2019pushing}. 
However, GNN-based approaches often encounter the challenge of over-squashing, where information is overly compressed when passed between long-distance atoms \cite{di2023over}. This compression may lead to the loss of crucial structural details, impairing the accuracy of molecular representations.

\citeauthor{liu2024kan} recently introduce the Kolmogorov-Arnold Network (KAN), which has been widely acknowledged for its high accuracy in small-scale AI + Science tasks. 
Unlike Multi-Layer Perceptrons (MLPs) \cite{cybenko1989approximation}, which use fixed activation functions on nodes (``neurons"), KAN employs learnable activation functions on edges (``weights"). This unique design greatly enhances KAN's performance, achieving 100 times greater accuracy than a four-layer MLP when solving partial differential equations (PDEs) with a two-layer structure \cite{liu2024kan}.
Additionally, KAN excels in various tasks, including image recognition \cite{azam2024suitability, wang2024spectralkan}, image generation \cite{li2024u}, and time series prediction tasks \cite{vaca2024kolmogorov, xu2024kolmogorov}. 
Despite its success, KAN has not yet become the standard on public molecular representation benchmarks.

In this paper, we observe that KAN's robust data-fitting ability and high accuracy in small-scale AI + Science tasks make it ideal for addressing the challenges of insufficient labeled data in molecular representation learning. 
Additionally, KAN's ability to effectively capture important molecular structure details helps to mitigate the limitations of GNN-based architectures, such as over-squashing.
Several studies \cite{kiamari2024gkan, bresson2024kagnns, zhang2024graphkan} have integrated KAN into graph-based tasks, highlighting its potential. However, these efforts often overlook the efficiency challenges of KAN. Moreover, there has been limited effort to adapt KAN specifically for molecular representation learning. 
Consequently, integrating KAN into standard GNNs for molecular representation learning remains an open problem.

To bridge this gap, we introduce a new class of GNNs, \textbf{GNN-SKAN}, and its augmented variant, \textbf{GNN-SKAN+}, tailored for molecular representation learning. 
Our key innovation lies in adapting KAN to molecular graphs by developing a variant that preserves KAN’s high accuracy in small-scale tasks while improving efficiency and flexibility in handling diverse molecular structures. 
Herein, we propose \textbf{SwallowKAN} (\textbf{SKAN}), which replaces B-splines with adaptive RBFs to achieve these goals. 
By combining the strengths of SKAN and GNNs, we have developed a robust and versatile approach to molecular representation learning. 
%
%
Specifically, we design a hybrid architecture that integrates GNN and SKAN, using a KAN-enhanced message-passing mechanism. To ensure versatility, SKAN is used as the update function to boost the expressiveness of the message-passing process. Additionally, we introduce GNN-SKAN+ which employs SKAN as the classifier to further reinforce model performance.
Notably, 
our approach holds three key benefits:
1) Superior Performance: GNN-SKAN and GNN-SKAN+ demonstrate superior prediction ability, robust generalization to unseen molecular scaffolds, and versatile transferability across different GNN architectures.
2) Efficiency: Our models achieve comparable or superior results to the SOTA self-supervised methods, while requiring less computational time and fewer parameters (Figure~\ref{fig:hiv_visualization}. 3) 
 Few-shot Learning Ability: GNN-SKAN shows great potential in few-shot learning scenarios, achieving an average improvement of 6.01\% across few-shot benchmarks.

The main contributions of this work are summarized as follows:
\begin{itemize}
    \item 
    To the best of our knowledge, we are the first to integrate KAN's exceptional approximation capabilities with GNNs' strengths in handling molecular graphs. This integration has led us to propose a new class of GNNs, termed GNN-SKAN, and its augmented variant, GNN-SKAN+, for molecular representation learning.
    \item 
    We propose SwallowKAN (SKAN), which introduces learnable RBFs as base function. This modification not only effectively addresses the slow speed of the original KAN but also significantly enhances the model's adaptability to diverse data distributions.
    \item We conduct a theoretical analysis of the parameter counts and computational complexity of KAN and SKAN. Our comparative analysis shows that SKAN exhibits a higher computational efficiency. 
    \item Our method achieves high accuracy and robust generalization on molecular property prediction tasks, surpassing or matching the SOTA models with lower time and memory requirements (Figure~\ref{fig:hiv_visualization}). It also excels in few-shot scenarios, with an average improvement of 6.01\% across 4 few-shot learning benchmarks (Section~\ref{section:experiment}).
\end{itemize}

\begin{figure}[htb]
    \centering
    \includegraphics[width=1.0\linewidth]{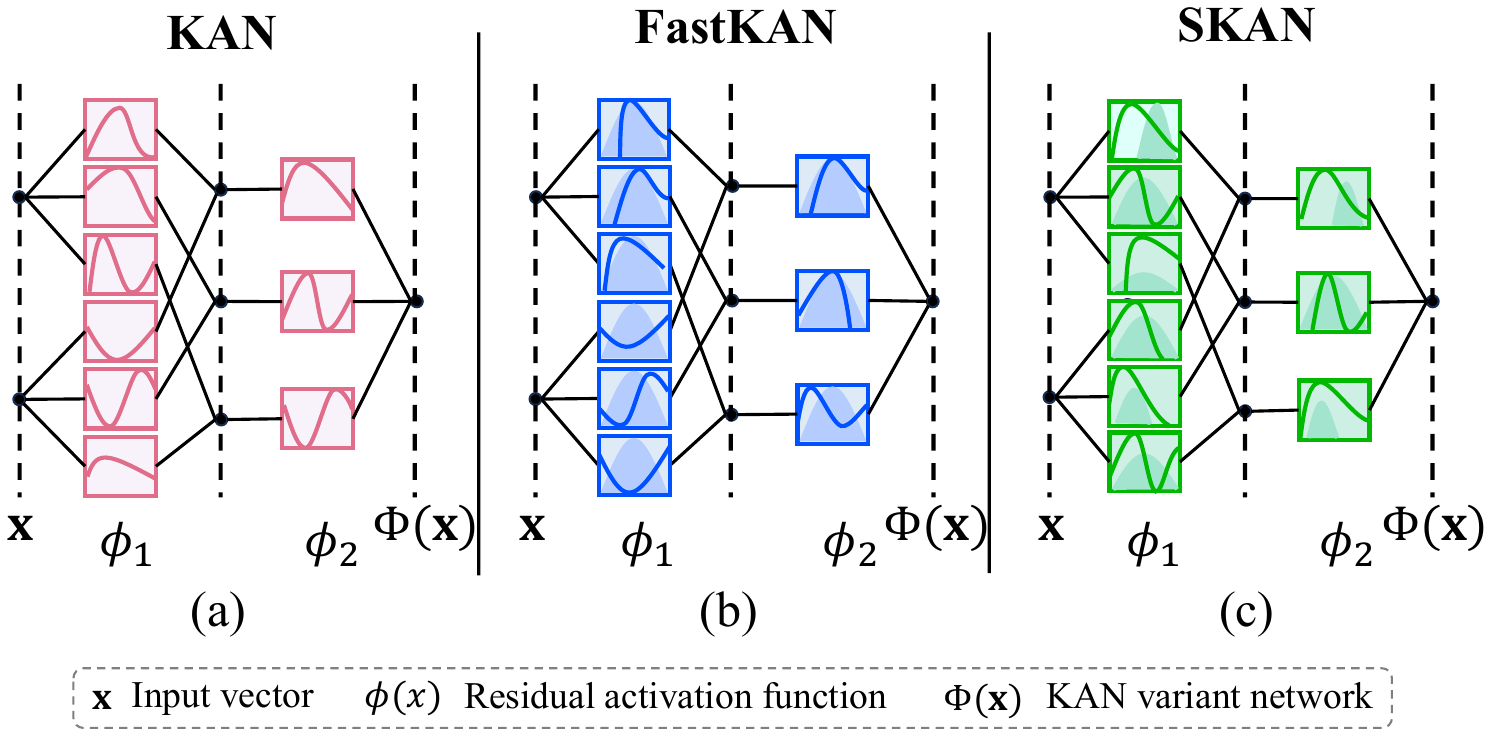}
    \caption{Comparison of three Kolmogorov-Arnold Network (KAN) variants, including (a) KAN \cite{liu2024kan} designed with B-splines, (b) FastKAN \cite{li2024kolmogorov} designed with RBFs, and (c) SwallowKAN (Our SKAN) designed with adaptable RBFs. The differences in base functions with each KAN variant are reflected in the shaded areas of each subfigure.}
    \label{fig:the_different_of_kan}
    \vspace{-15pt}
\end{figure}

\section{Related Work}

\paragraph{Molecular Representation Learning.}
Existing molecular representation learning approaches can be broadly classified into three categories: 1) GNN-based approaches: Methods like MolCLR \cite{wang2022molecular} and KANO \cite{fang2023knowledge} leverage contrastive learning framework, while MGSSL \cite{zhang2021motif} and GraphMAE \cite{hou2022graphmae} employ generative pre-training strategies. Although these methods effectively learn molecular representations, they face the challenge of over-squashing \cite{black2023understanding}. 
2) Transformer-based approaches: Models such as MolBERT \cite{fabian2020molecular}, ChemBERTa \cite{chithrananda2020chemberta}, ChemBERTa-2 \cite{ahmad2022chemberta}, SELFormer \cite{yuksel2023selformer}, and MolGen \cite{fang2024domainagnostic} enhance molecular representation learning with the self-attention mechanism, but they suffer from the quadratic complexity \cite{vaswani2017attention}. 3) Combination of both (Graph Transformers): These models, including Graphormer \cite{ying2021transformers}, GPS \cite{rampavsek2022recipe}, Grover \cite{rong2020self}, and Uni-Mol \cite{zhou2023unimol}, which combine message-passing neural networks with Transformer to enhance expressive power. However, they also suffer from high computational complexity \cite{keles2023computational}. 
Unlike previous methods, we uniquely integrate GNNs with SKAN, a proposed KAN variant, to overcome the challenges present in existing molecular representation learning approaches.

\paragraph{Kolmogorov-Arnold Networks (KANs).}
KAN is well-regarded for its ability to approximate complex functions using learnable activation functions on edges \cite{liu2024kan}. 
Recent work has extended KAN to the graph-structured domain. For instance, GKAN \cite{kiamari2024gkan, de2024kolmogorov}, KAGNNs \cite{bresson2024kagnns}, and GraphKAN \cite{zhang2024graphkan} integrate KAN into GNNs to enhance performance. 
 However, these approaches do not address KAN's efficiency issues. For example, GKAN incurs computational costs approximately 100 times higher than the standard GNNs on the Cora dataset \cite{de2024kolmogorov}, significantly hindering the broader adoption of KAN in graph-based applications.
To improve the efficiency of KAN, several approaches have been proposed to replace B-splines with simpler base functions \cite{ta2024bsrbf}. Notably, FastKAN \cite{li2024kolmogorov} addresses this issue by replacing B-splines with RBFs \cite{wu2012using}, and GP-KAN \cite{chen2024gaussian} employs Gaussian Processes (GPs) \cite{seeger2004gaussian} as an alternative, particularly for image classification tasks. 
Unlike these approaches, we introduce SKAN, a new KAN variant that incorporates adaptable RBFs specifically designed for efficient molecular representation learning. The differences between KAN, FastKAN and our SKAN are shown in Figure \ref{fig:the_different_of_kan}.

\section{The proposed approach: GNN-SKAN}
\subsection{Overview}
\begin{figure}
    \centering
    \includegraphics[width=1.\linewidth]{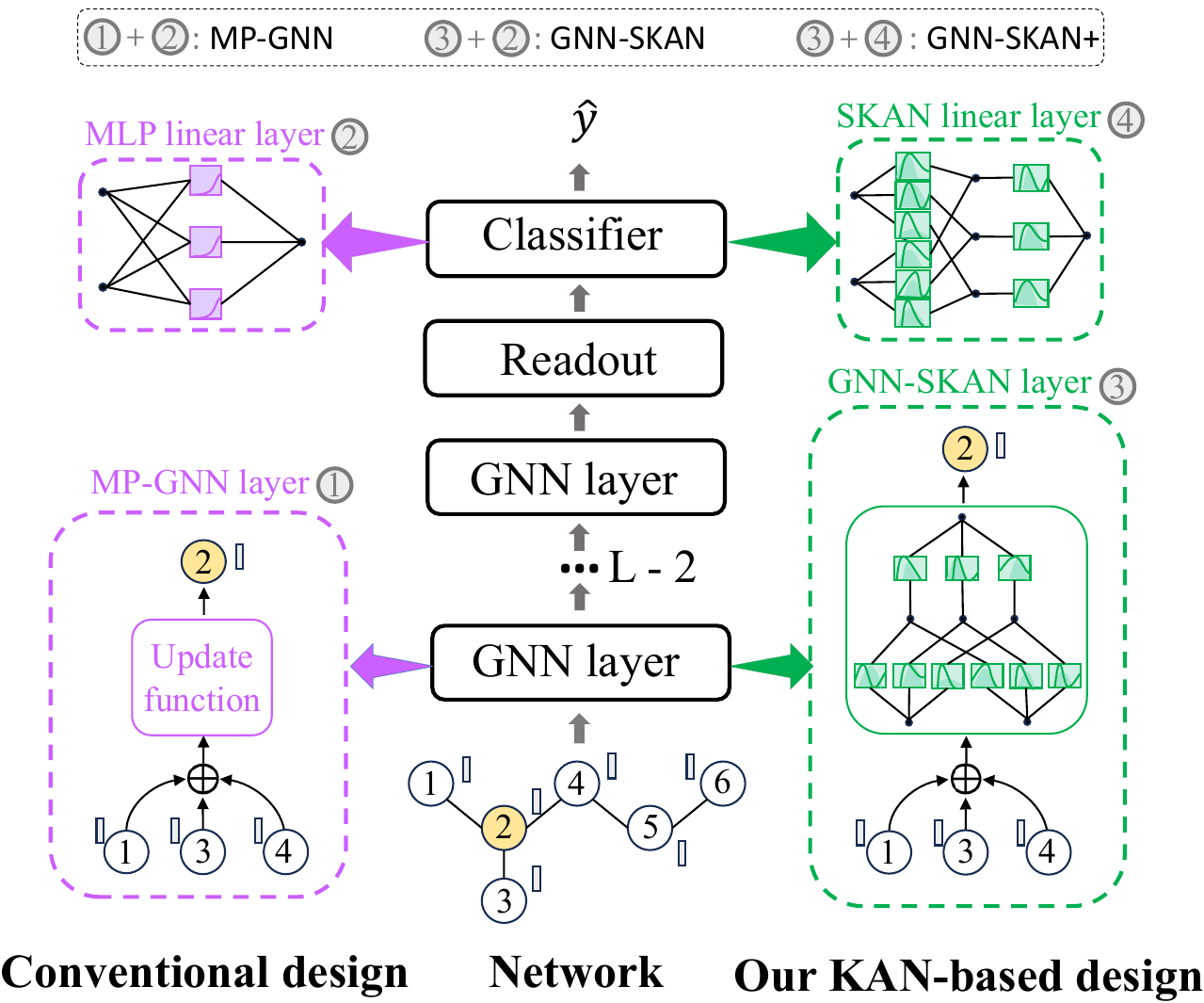}
    \caption{
    Comparison of the network architecture between MP-GNNs and our GNN-SKAN and GNN-SKAN+. 
    }
    \label{fig:method}
    \vspace{-10pt}
\end{figure}

The framework of GNN-SKAN is illustrated in Figure~\ref{fig:method}. This section focuses on the details of the implementation of our method. These details are crucial for ensuring the speed and reliability of the network.

\paragraph{Notation.}
Let $G=(\mathcal{V}, \mathcal{E})$ be a molecular graph, where $\mathcal{V}$ represents the set of nodes and $\mathcal{E}$ represents the set of edges. 
The number of nodes is $N = |\mathcal{V}|$. 
Let $\mathbf{H} = \{\mathbf{h}_v^{(0)}| v \in \mathcal{V}\}$ denote the initial features of nodes, and $\mathbf{E} = \{\mathbf{b}_{uv} | (u, v) \in \mathcal{E}\}$ denote the edge features. 
Let $\mathbf{z}$ represent the graph-level representation, $\hat{y}$ represent the graph-level prediction, and $y$ represent the molecular property label (i.e., the ground-truth label). For the RBF, $c_w$ and $bw_w$ are the learnable center and bandwidth, respectively.

\subsection{Preliminary}
\paragraph{Kolmogorov-Arnold Representation Theorem (KART).} KART \cite{arnold2009representation, kolmogorov1957presentation, braun2009constructive} states that any multivariate continuous function defined on a bounded domain can be expressed as a finite composition of continuous single-variable functions and addition operations. Formally, for a smooth function $f:[0,1]^n \rightarrow \mathbb{R}$ with $n$ variables $\mathbf{x}=x_1, x_2, \ldots, x_n$, the theorem guarantees the existence of the following representation: 
\begin{equation}
\label{eq:KART}
f\left(x_1, \ldots, x_n\right)=\sum_{q=1}^{2 n+1} \Phi_q\left(\sum_{p=1}^n \varphi_{q, p}\left(x_p\right)\right) \text {, }
\end{equation}
where $\Phi_q:\mathbb{R} \rightarrow \mathbb{R}$ and $\varphi_{q, p}:[0,1] \rightarrow \mathbb{R}$.

\paragraph{The Design of KAN.}
 Unlike the simple 2-layer KAN described by Eq.~\ref{eq:KART}, which follows a $[n, 2n+1, 1]$ structure, KAN can be generalized to arbitrary widths and depths. 
 Specifically, KAN consists of $L$ layers and generates outputs based on the input vector $\mathbf{x} \in \mathbb{R}^{n}$: 
\begin{equation}
\operatorname{KAN}(\mathbf{x})=\left(\Phi_{L-1} \circ \Phi_{L-2} \circ \cdots \circ \Phi_1 \circ \Phi_0\right) \mathbf{x} \text {, }
\end{equation}
where $\Phi_l$ represents the function matrix of the $l^{th}$ layer. Each layer of KAN, defined by the function matrix $\Phi_l$, is formed as follows:
\begin{equation}
\fontsize{9}{11}\selectfont
\mathbf{x}_{l+1}=\underbrace{\left(\begin{array}{cccc}
\phi_{l, 1,1}(\cdot) & \phi_{l, 1,2}(\cdot) & \cdots & \phi_{l, 1, n_l}(\cdot) \\
\phi_{l, 2,1}(\cdot) & \phi_{l, 2,2}(\cdot) & \cdots & \phi_{l, 2, n_l}(\cdot) \\
\vdots & \vdots & & \vdots \\
\phi_{l, n_{l+1}, 1}(\cdot) & \phi_{l, n_{l+1}, 2}(\cdot) & \cdots & \phi_{l, n_{l+1}, n_l}(\cdot)
\end{array}\right)}_{\boldsymbol{\Phi}_l} \mathbf{x}_l \text {, }
\end{equation}
where $\phi_{l, j, i}$ denotes the activation function connecting neuron $i$ in layer $l$ to neuron $j$ in layer $l+1$. The network utilizes $n_l \times n_{l+1}$ such activation functions between each layer. 

KAN enhances its learning capability by utilizing residual activation functions $\phi(x)$ that integrate a basis function and a spline function:
\begin{equation}
\phi(x)=w_b b(x)+w_s \operatorname{ Spline }(x) \text {, }
\end{equation}
where $b(x)$ is the $silu(x)$ function, and $\operatorname{Spline}(x)$ is a linear combination of B-splines \cite{unser1993b}. The parameters $w_b$ and $w_s$ are learnable.

%
%

\subsection{Architecture}
Molecular representation learning with MP-GNNs generally involves three key steps: Aggregation, Update, and Readout.  
In this work, we generalize KAN to molecular graphs for molecular representation learning. 
To address KAN's limitations in computational efficiency and adaptability, we introduce SwallowKAN (SKAN), a variant of KAN that employs learnable RBFs as activation functions. SKAN adjusts its parameters adaptively based on the data distribution, enabling effective modeling of molecules with diverse scaffolds. Building on the strengths of SKAN, 
we propose a new class of GNNs, GNN-SKAN, and its advanced version, GNN-SKAN+, which 
use a KAN-enhanced message-passing mechanism to improve molecular representation learning.
Specifically, SKAN is employed as the update function, allowing seamless integration into any MP-GNN. For GNN-SKAN+, SKAN also serves as the classifier to further enhance performance.
The complete algorithm of our approach is shown in Algorithm~\ref{algo}.

\paragraph{Step 1: Aggregation.}
In a molecular graph $G=(\mathcal{V}, \mathcal{E})$, let $\mathbf{h}^{(0)} \in \mathbb{R}^n$ represent the initial node features and $\mathbf{b} \in \mathbb{R}^n$ represent the initial edge features. At the $d^{th}$ layer, GNN utilizes an aggregation function to combine information from the neighboring nodes of $v$, forming a new message representation $\mathbf{m}_v^{(d)} \in \mathbb{R}^n$:
\begin{align}\label{eq:aggregate}
\mathbf{m}_v^{(d)} =& \operatorname{AGGREGATE}^{(d)}\left(\left\{\left(\mathbf{h}_v^{(d-1)}, \mathbf{h}_u^{(d-1)}, \mathbf{b}_{vu}\right) \right. \right. \nonumber\\
&\left. \left. \mid u \in \mathcal{N}(v)\right\}\right)
\end{align}
where $u \in \mathcal{N}(v)$ denotes the set of neighboring nodes of node $v$. 

\paragraph{Step 2: Update.}
After computing the aggregated message $\mathbf{m}_v^{(d)}$ for node $v$, the message is used to update the representation of node $v$ at layer $d$:
\begin{equation}
\label{eq:update_function}
\mathbf{h}_v^{(d)}=\operatorname{UPDATE}^{(d)}\left(\mathbf{h}_v^{(d-1)}, \mathbf{m}_v^{(d)}\right) \text{.}
\end{equation}

To efficiently adapt to various data distributions, we design an augmented KAN, SwallowKAN (SKAN), with learnable RBFs to replace the original update function. The data center $c_{w}$ and the bandwidth $bw_w$ efficiently adapt to the data distribution. The update function can be rewritten as follows:
\begin{equation}\label{eq:update}
\fontsize{9}{10}\selectfont
\mathbf{h}_v^{(d)}=\operatorname{SKAN}^{(d)}\left(\left(1+\epsilon^{(d)}\right) \cdot \mathbf{h}_v^{(d-1)}+\mathbf{m}_{v}^{(d)}\right) \text{,}
\end{equation}
where $\epsilon^{(d)}$ is a parameter.  Specifically, let $\mathbf{x} = \left(1+\epsilon^{(d)}\right) \cdot \mathbf{h}_v^{(d-1)}+\mathbf{m}_v^{(d)}$. The residual activation function of SKAN at the $d^{th}$ layer is defined as:
\begin{equation}\label{}
\fontsize{9.5}{10}\selectfont
\phi\left(x\right)=w_b b\left(x\right)+w_r \exp \left(-\frac{1}{2}\left(\frac{x-c_w}{bw_w}\right)^2\right)\footnote{For ease of notation, each variable in this equation represent a single variable without a subscript.}  \text{.}
\end{equation} 

SKAN's adaptive parameter adjustment allows it to effectively handle the varied nature of molecular structures, leading to a more robust and adaptable model with improved generalization across diverse molecular datasets.

\paragraph{Step 3: Readout.}
After performing Steps 1 and 2 for $D$ iterations, we use the MEAN function as the readout mechanism to obtain the molecular representation. The readout operation is defined as:
\begin{equation}\label{eq:readout}
\mathbf{g}=\operatorname{READOUT}\left(\left\{\mathbf{h}_v^{(D)} \mid v \in \mathcal{V}\right\}\right) \text{,}
\end{equation}
where $\mathbf{g} \in \mathbb{R}^n$ represents molecular representation. 
\paragraph{Step 4: Prediction.} Next, we leverage an MLP to obtain the final prediction $\hat{y}$:
\begin{equation}\label{eq:predict}
\hat{y}=\operatorname{MLP}\left(\mathbf{g}\right) \in  \mathbb{R}^{n_t} \text{,}
\end{equation}
where $n_t$ denotes the number of tasks in the molecular property classification or regression datasets. For the advanced version, GNN-SKAN+, we utilize a 2-layer SKAN as the classifier:
\begin{equation}\label{eq:predict+}
\hat{y}=\operatorname{SKAN}\left(\mathbf{g}\right) \in \mathbb{R}^{n_t} \text{.}
\end{equation}

\vspace{-3mm}
\begin{algorithm}[htb]
\caption{GNN-SKAN and GNN-SKAN+ for Molecular representation Learning}
\label{algo}
\begin{algorithmic}[1]
\STATE \textbf{Input:} Molecular graph $G = (\mathcal{V}, \mathcal{E})$ with initial node features $\mathbf{H} = \{\mathbf{h}_v^{(0)}| v \in \mathcal{V}\}$ and edge features $\mathbf{E} = \{\mathbf{b}_{uv} | (u, v) \in \mathcal{E}\}$, number of layers $D$
\STATE \textbf{Output:} Molecular property prediction $\hat{y}$

\STATE Initialize node representations: $\mathbf{h}_v^{(0)}$ for all $v \in V$

\FOR{each iteration $d = 1$ to $D$}
    \FOR{each node $v \in V$}
        \STATE Aggregate messages from neighbors by Eq.~\ref{eq:aggregate}
    \ENDFOR
    
    \FOR{each node $v \in V$}
        \STATE Update node representation by Eq.~\ref{eq:update}
    \ENDFOR
\ENDFOR

\STATE Perform readout for molecular representation by Eq.~\ref{eq:readout}
\STATE Predict molecular property by Eq.~\ref{eq:predict} (for GNN-SKAN) or Eq.~\ref{eq:predict+} (for GNN-SKAN+)
\STATE \textbf{return} $\hat{y}$
\end{algorithmic}
\end{algorithm}

\subsection{The efficiency of SKAN}
For a SKAN network, we denote $L$ as the number of layers in the network, 
$N$ as the number of nodes in each layer,
and $M$ as the number of RBFs.

\begin{table*}[htb]
\centering
\begin{tabular}{lcccccc}
\toprule[1pt]
\multicolumn{1}{l}{Model} & \multicolumn{1}{c}{BBBP $\uparrow$}         & \multicolumn{1}{c}{Tox21 $\uparrow$}      & \multicolumn{1}{c}{ToxCast $\uparrow$}    & \multicolumn{1}{c}{SIDER $\uparrow$}      & \multicolumn{1}{c}{HIV $\uparrow$}        & \multicolumn{1}{c}{BACE $\uparrow$}       \\ 
\midrule
GCN                       & 0.622 $\pm$ 0.037 & 0.738 $\pm$ 0.003 & 0.624 $\pm$ 0.006 & 0.611 $\pm$ 0.012 & 0.748 $\pm$ 0.007 & 0.696 $\pm$ 0.048 \\
GCN-GKAN 
& - & 0.742 $\pm$ 0.005  &  - & - & - &  0.732 $\pm$ 0.013
\\
GCN-SKAN              & 0.652 $\pm$ 0.009 & 0.737 $\pm$ 0.003 & 0.632 $\pm$ 0.006 & 0.603 $\pm$ 0.002 & 0.757 $\pm$ 0.021 & \textbf{0.754 $\pm$ 0.023} \\
GCN-SKAN+             & \textbf{0.676 $\pm$ 0.014} & \textbf{0.747 $\pm$ 0.005} & \textbf{0.643 $\pm$ 0.004} & \textbf{0.614 $\pm$ 0.005} & \textbf{0.786 $\pm$ 0.015} & 0.747 $\pm$ 0.009 \\ 
\midrule
GAT                       & 0.640 $\pm$ 0.032 & 0.732 $\pm$ 0.003 & 0.635 $\pm$ 0.012 & 0.578 $\pm$ 0.017 & 0.751 $\pm$ 0.010 & 0.653 $\pm$ 0.016 \\
GAT-GKAN 
& - &  0.725 $\pm$ 0.008 &  - & - & - & 0.712 $\pm$ 0.014
\\
GAT-SKAN              & 0.649 $\pm$ 0.007 & 0.723 $\pm$ 0.005 & 0.633 $\pm$ 0.003 & 0.603 $\pm$ 0.004 & 0.746 $\pm$ 0.016 & \textbf{0.755 $\pm$ 0.028} \\
GAT-SKAN+             & \textbf{0.688 $\pm$ 0.015} & \textbf{0.748 $\pm$ 0.004} & \textbf{0.640 $\pm$ 0.004} & \textbf{0.620 $\pm$ 0.002} & \textbf{0.784 $\pm$ 0.004} & 0.731 $\pm$ 0.060 \\ 
\midrule
GINE                       & 0.644 $\pm$ 0.018 & 0.733 $\pm$ 0.009 & 0.612 $\pm$ 0.009 & 0.596 $\pm$ 0.003 & 0.744 $\pm$ 0.014 & 0.612 $\pm$ 0.056 \\
GINE-GKAN 
& - & 0.734 $\pm$ 0.011  &  - & - & - &  0.706 $\pm$ 0.018
\\
GINE-SKAN              & 0.652 $\pm$ 0.012 & 0.738 $\pm$ 0.008 & 0.627 $\pm$ 0.005 & 0.586 $\pm$ 0.004 & 0.764 $\pm$ 0.008 & \textbf{0.768 $\pm$ 0.011} \\
GINE-SKAN+             & \textbf{0.688 $\pm$ 0.016} & \textbf{0.750 $\pm$ 0.002} & \textbf{0.630 $\pm$ 0.005} & \textbf{0.613 $\pm$ 0.004} & \textbf{0.785 $\pm$ 0.009} & 0.762 $\pm$ 0.027 \\ 
\bottomrule[1pt]
\end{tabular}
\caption{Comparison with the base MP-GNNs on six molecular classification benchmarks. The best results are in bold. The mean and standard deviation of ROC-AUC over five independent test runs are reported. `-' indicates out of memory.}
\label{table:classification_base_gnn}
\end{table*}

\paragraph{Parameter Counts.}
In the original KAN, the total number of parameters is $O(LN^2(G+k))$, where $G$ represents the number of grid points and $k$ (typically $k = 3$) is the order of each spline. 
Since \( k \) is much smaller than \( G \) \cite{liu2024kan} and its impact on the total number of parameters is usually negligible, \( O(N^2L(G + k)) \) simplifies to \( O(N^2LG) \). 
In comparison, SKAN has a total of $O(LN^2M)$ parameters, where $M \leq 8 $, which is smaller than \(G\). Thus, SKAN requires fewer parameters than the original KAN.

\paragraph{Computational Complexity.} The computational complexity of the original KAN is \(O((LN^2)k)\), where $k$ is the order of the spline. In contrast, SKAN's computational complexity is \(O(LN^2)\) because each RBF has a complexity of \(O(1)\). 

Overall, SKAN demonstrates clear advantages over the original KAN in terms of parameter efficiency and computational complexity, leading to greater scalability. Notably, with its efficiency and high accuracy, our GNN-SKAN and GNN-SKAN+ models, despite having two layers, outperform the 5-layer base GNNs  and show comparable or superior performance to self-supervised SOTA methods while achieving higher efficiency.

\section{Experiments}\label{section:experiment}
\paragraph{Molecular Benchmark Datasets.} 
We evaluate our GNN-SKAN and GNN-SKAN+ across a wide range of public molecular property prediction benchmarks. The evaluation includes three types: 1) \textbf{Molecular Classification Datasets}: BBBP, Tox21, ToxCast, SIDER, HIV, and BACE; 2) \textbf{Molecular Regression Datasets}: Lipo, FreeSolv, Esol, QM7, QM8, and QM9; and 3) \textbf{Few-shot Learning Datasets}: Tox21, SIDER, MUV, and ToxCast from PAR \cite{wang2021propertyaware}. 

\paragraph{Experimental Setup.}

Following the experimental protocols established by \citet{Hu*2020Strategies} and \citet{ hou2022graphmae}, we use ROC-AUC as the evaluation metric for classification tasks and mean absolute error (MAE) for regression tasks.  The results are reported as the mean and standard deviation across five independent test runs. Our model employs a 2-layer GNN-SKAN as the encoder with 256 hidden units, paired with an MLP as the classifier. All base models use a 5-layer architecture. Additionally, we introduce GNN-SKAN+ with a 2-layer SKAN classifier. The learning rate ranges from 0.01 to 0.001 and the number of adaptive RBFs ranges from 3 to 8. 
All experiments are conducted on a single NVIDIA RTX A6000 GPU. 

\subsection{Classification Tasks}

\begin{table*}[htb]
\begin{tabular}{lllllll}
\toprule[1pt]
\multicolumn{1}{l}{Model} & \multicolumn{1}{c}{Lipo $\downarrow$}       & \multicolumn{1}{c}{FreeSolv $\downarrow$}   & \multicolumn{1}{c}{Esol $\downarrow$}       & \multicolumn{1}{c}{QM7 $\downarrow$}          & \multicolumn{1}{c}{QM8 $\downarrow$}        & \multicolumn{1}{c}{QM9 $\downarrow$}        \\ 
\midrule
GCN                         & 0.809 $\pm$ 0.053 & 2.891 $\pm$ 0.249 & \textbf{1.241 $\pm$ 0.086} & 79.393 $\pm$ 2.789  & 0.025 $\pm$ 0.001 & 0.311 $\pm$ 0.008 \\
GCN-SKAN              & 0.921 $\pm$ 0.064 & \textbf{1.888 $\pm$ 0.043} & 1.261 $\pm$ 0.049 & 79.443 $\pm$ 0.649  & \textbf{0.023 $\pm$ 0.000} & 0.266 $\pm$ 0.004                             \\
GCN-SKAN+             & \textbf{0.771 $\pm$ 0.015} & 1.893 $\pm$ 0.052 & 1.410 $\pm$ 0.074 & \textbf{77.512 $\pm$ 0.987}  & \textbf{0.023 $\pm$ 0.000} & \textbf{0.260 $\pm$ 0.007}                               \\ 
\midrule
GAT                       & 0.800 $\pm$ 0.025 & 2.360 $\pm$ 0.501 & \textbf{1.345 $\pm$ 0.095} & 112.989 $\pm$ 4.884 & 0.025 $\pm$ 0.001   & 0.380 $\pm$ 0.008 \\
GAT-SKAN              & 0.824 $\pm$ 0.015 & \textbf{1.935 $\pm$ 0.123} & 1.348 $\pm$ 0.089 & 105.494 $\pm$ 4.758 & 0.026 $\pm$ 0.001 & 0.374 $\pm$ 0.007                               \\
GAT-SKAN+             & \textbf{0.777 $\pm$ 0.020} & 2.077 $\pm$ 0.174 & 1.347 $\pm$ 0.079 & \textbf{100.664 $\pm$ 3.939} &\textbf{0.024 $\pm$ 0.000} & \textbf{0.371 $\pm$ 0.010}                        \\  
\midrule         
GINE                      & 1.158 $\pm$ 0.227 & 2.086 $\pm$ 0.092 & 1.275 $\pm$ 0.077 & 94.559 $\pm$ 23.708 & 0.034 $\pm$ 0.013 & 0.347 $\pm$ 0.013 
\\
GINE-SKAN             & 0.934 $\pm$ 0.121 & 1.722 $\pm$ 0.125 & \textbf{1.098 $\pm$ 0.018} & 78.200 $\pm$ 1.536  & \textbf{0.022 $\pm$ 0.000} &  \textbf{0.250 $\pm$ 0.004}   \\
GINE-SKAN+            & \textbf{0.782 $\pm$ 0.002} & \textbf{1.715 $\pm$ 0.130} & 1.175 $\pm$ 0.022 & \textbf{77.315 $\pm$ 2.114}  & 0.024 $\pm$ 0.001 & 0.254 $\pm$ 0.002        \\ 
\bottomrule[1pt]
\end{tabular}
\caption{Comparison with standard GNNs on six molecular regression benchmarks. The best results are indicated in bold. The mean and standard deviation of the test mean absolute error on five independent runs are reported.}
\label{table:regression_task}
\end{table*}

\begin{table*}[]
\centering
\begin{tabular}{llcccc}
\toprule[1pt]
\multicolumn{1}{l}{Model}& \multicolumn{1}{l}{\#Param.} & \multicolumn{1}{c}{Tox21 (10-shot)} & \multicolumn{1}{c}{SIDER (10-shot)} & \multicolumn{1}{c}{MUV (10-shot)}   & \multicolumn{1}{c}{ToxCast (10-shot)} \\
\midrule
TPN                       & 2,076.73K 
& 0.795 $\pm$ 0.002 & 0.675 $\pm$ 0.003 & 0.783 $\pm$ 0.007 &  0.708 $\pm$ 0.006                                \\
TPN-SKAN                 & 
982.51K 
 & \textbf{0.834 $\pm$ 0.002} & \textbf{0.799 $\pm$ 0.004} & \textbf{0.835 $\pm$ 0.005} &   \textbf{0.755 $\pm$ 0.001}                               \\ 
\midrule
MAML                     & 
1,814.94K 
& 0.762 $\pm$ 0.001 & 0.707 $\pm$ 0.001 & \textbf{0.814 $\pm$ 0.002} &  0.699 $\pm$ 0.017                                \\
MAML-SKAN               & 
736.55K
& \textbf{0.809 $\pm$ 0.002} & \textbf{0.828 $\pm$ 0.001} & 0.784 $\pm$ 0.002 &  \textbf{0.703 $\pm$ 0.015}                                \\ 
\midrule
ProtoNet             &  1,814.36K 
& 0.634 $\pm$ 0.001 & \textbf{0.592 $\pm$ 0.001} & \textbf{0.719 $\pm$ 0.001} &    0.637 $\pm$ 0.015                              \\
ProtoNet-SKAN        & 
732.63K 
& \textbf{0.705 $\pm$ 0.002} & 0.589 $\pm$ 0.001 & 0.705 $\pm$ 0.002 &              \textbf{0.639 $\pm$ 0.012}                    \\ 
\midrule 
PAR                       & 
2,377.91K 
& 0.804 $\pm$ 0.002 & 0.680 $\pm$ 0.001 & 0.780 $\pm$ 0.001 &  0.698 $\pm$ 0.004                                \\
PAR-SKAN             & 1,124.70K 
& \textbf{0.830 $\pm$ 0.003}          & \textbf{0.829 $\pm$ 0.001} & \textbf{0.831 $\pm$ 0.001} &   \textbf{0.751 $\pm$ 0.002}                               \\ 
\bottomrule 
\end{tabular}
\caption{Comparison of common few-shot learning methods on four few-shot molecular property prediction benchmarks. The best results are in bold. Mean and standard deviation of test ROC-AUC across five independent runs.}
\label{table:few-shot}
\end{table*}

\paragraph{Comparison with Base MP-GNNs.}
Table~\ref{table:classification_base_gnn} summarizes the performance of three base MP-GNNs---GCN \cite{kipf2017semisupervised},  GAT \cite{velivckovic2018graph}, and GINE \cite{Hu*2020Strategies}---with their augmented versions, including GKAN~\cite{kiamari2024gkan}, SKAN, and SKAN+, across six classification benchmarks. Given the architectural similarities among GKAN, GraphKAN \cite{zhang2024graphkan}, and KAGNNs \cite{bresson2024kagnns}, 
we extend the recently proposed GKAN as a baseline to other MP-GNNs.
In Table \ref{table:classification_base_gnn},
our augmented models, GNN-SKAN and GNN-SKAN+, consistently outperform the base MP-GNNs and GNN-GKAN across most benchmarks. 
Notably, GCN-SKAN improves by 4.82\% over GCN, and GCN-SKAN+ improves by 8.68\% over GCN on the BBBP dataset. 
Using scaffold splitting \cite{mayr2018large, yang2019analyzing} to evaluate generalization, the augmented models show solid performance on new scaffolds.
For instance, GINE-SKAN outperforms GINE by 25.49\% on the BACE dataset, demonstrating robust generalization.  These results indicate that the augmented models effectively capture structural information of molecules and adapt to diverse molecular scaffolds. 
Additionally, our augmented models demonstrate outstanding performance across different base MP-GNNs, highlighting their transferability and versatility.


\paragraph{Comparison with SOTAs.} Figure~\ref{fig:hiv_visualization} compares the best results of our augmented models from Table~\ref{table:classification_base_gnn} with the SOTAs on the HIV dataset. The baselines are divided into two categories: 1) Supervised methods: GCN \cite{kipf2017semisupervised}, GAT \cite{velivckovic2018graph}, GINE \cite{Hu*2020Strategies}, and CMPNN \cite{song2020communicative}; 2) Self-supervised methods: MolCLR \cite{wang2022molecular}, MGSSL \cite{zhang2021motif}, and GraphMAE \cite{hou2022graphmae}. Additional comparisons with Graph Transformer methods are provided in Appendix B.
In Figure \ref{fig:hiv_visualization}, our models, GNN-SKAN and GNN-SKAN+, consistently outperform or match the SOTA methods.
They also demonstrate a significantly reduction in the number of parameters---approximately 563K and 565K, respectively--- compared to the models like GraphMAE ($\sim$1363K) and MGSSL ($\sim$1858K), while maintaining comparable performance.
Moreover, GNN-SKAN and GNN-SKAN+ show a faster training speeds per epoch. In contrast, models with more parameters, such as GraphMAE and MolCLR, require significantly longer training times. This indicates that our models can complete training more quickly, improving efficiency for practical applications.

\subsection{Regression Tasks}
Table~\ref{table:regression_task} presents a comparison between the base MP-GNNs and their augmented counterparts, GNN-SKAN and GNN-SKAN+, across six regression benchmarks in physical chemistry and quantum mechanics. 
The augmented models consistently outperform their base counterparts across most datasets, highlighting the effectiveness of SKAN.
Remarkable advancements are observed in the Lipo and FreeSolv datasets. 
Specifically, GCN-SKAN+ and GINE-SKAN+ improve by 4.70\% and 32.47\% on the Lipo dataset and by 34.52\% and 17.78\% on the FreeSolv dataset. 
In summary, these findings underscore the potential of GNN-SKAN and GNN-SKAN+ to enhance the accuracy and reliability of GNNs in molecular regression tasks.

\subsection{Few-shot Learning Tasks}
Table~\ref{table:few-shot} demonstrates the effectiveness of SKAN in addressing the few-shot learning problem. We select four common few-shot baselines: TPN \cite{liu2018learning}, MAML \cite{finn2017model}, ProtoNet \cite{snell2017prototypical}, and PAR \cite{wang2021property}. All experiments are conducted in a 2-way 10-shot setting. 
We do not include 1-shot experiments, as they are less meaningful in drug discovery. 
In Table~\ref{table:few-shot}, the augmented models consistently outperform their base counterparts across most datasets, with average improvements of 9.14\% for TPN-SKAN, 5.04\% for MAML-SKAN, 2.26\% for ProtoNet-SKAN, and 7.59\% for PAR-SKAN. 
These augmentations with the SKAN architecture result in higher ROC-AUC values, improving the accuracy and reliability of few-shot molecular property prediction methods.

\subsection{Visualization}
\begin{figure}[htb]
    \centering
    \includegraphics[width=0.9\linewidth]{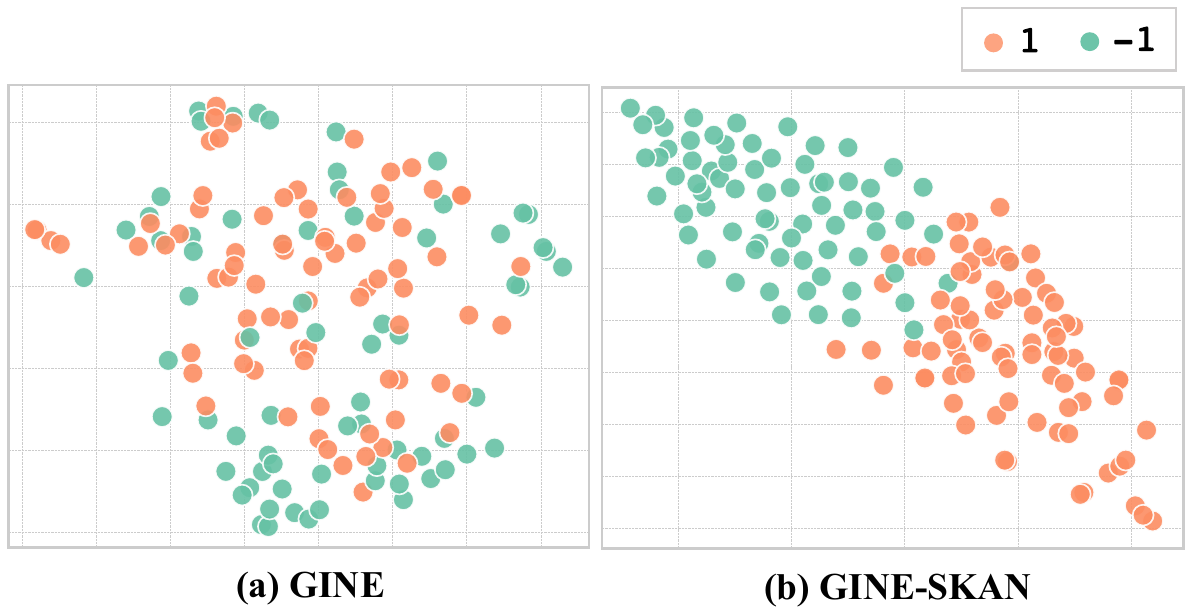}
    \caption{t-SNE visualization of molecular representations on the BACE dataset, extracted from (a) GINE and (b) GINE-SKAN. The green and orange dots represent molecules with labels -1 and 1, respectively.}
\label{fig:visualization}
\end{figure}

Figure~\ref{fig:visualization} shows the t-SNE visualization of molecular representations learned by GINE and GINE-SKAN on the BACE dataset. In the left plot, GINE appears to struggle with over-squashing, potentially losing key molecular structural details and resulting in overly similar representations. 
In contrast, the right plot, which displays the results from GINE-SKAN+, shows a clearer separation between the green dots (negative samples) and orange dots (positive samples). This improvement suggests that SKAN enhances the model's ability to capture and differentiate molecular structures, effectively alleviating the over-squashing issues common in standard GNNs.

\begin{figure}[htb]
    \centering
    \includegraphics[width=0.7\linewidth]{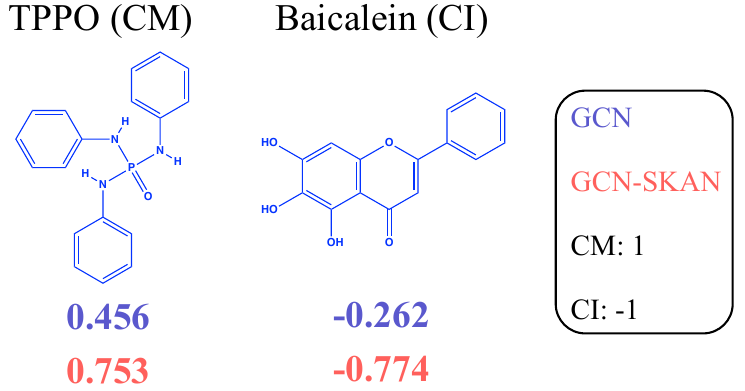}
    \caption{Predicted scores for TPPO (CM) and Baicalein (CI) using \textcolor{purplecolor}{GCN} and \textcolor{redcolor}{GCN-SKAN}. The CM class is labeled as 1, and the CI class as -1.}
    \label{fig:vis_2_mol}
\end{figure}

Additionally, the examples in Figure~\ref{fig:vis_2_mol} illustrate that GCN-SKAN exhibits higher confidence in its predictions for both TPPO and Baicalein on the HIV dataset. This further indicates that SKAN improves the model's generalization ability across diverse molecular structures, thereby improving overall predictive performance.

\subsection{Ablation Study}
We conduct ablation studies to evaluate the design choices for each component of the architecture.
 First, we examine the effectiveness of SKAN and its integration within GNNs. 

\begin{figure}[htb]
\centering
\includegraphics[width=1.0\linewidth]{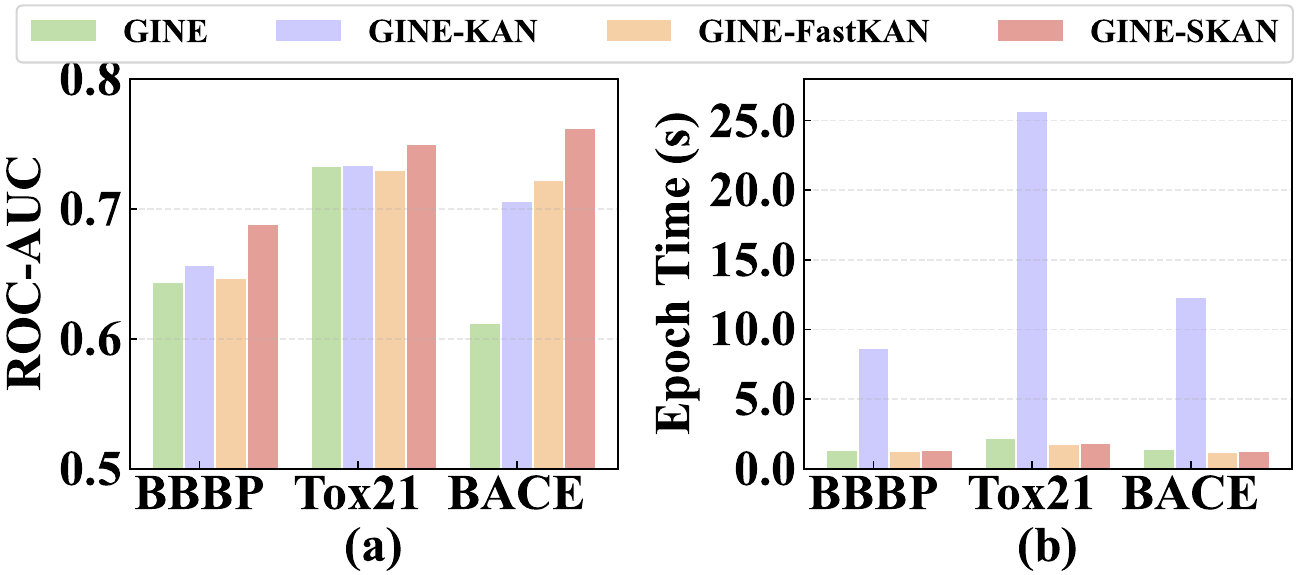}
    \caption{The comparison of (a) performance and (b) time cost between GINE and its augmented versions, including Original KAN, FastKAN, SKAN, across three datasets (BBBP, Tox21, BACE).}
    \label{fig:SKAN_efficiency}
    \vspace{-10pt}
\end{figure}

\paragraph{The Effectiveness of SKAN.} 
Figure~\ref{fig:SKAN_efficiency} (a) compares the performance of GINE and its KAN variants across three datasets. GINE-SKAN shows notably superior performance over GINE and other variants (GINE-KAN and GINE-FastKAN). The adaptive bandwidths and centers in SKAN enhance its ability to adapt to data distribution, thereby improving flexibility.
Figure~\ref{fig:SKAN_efficiency} (b) presents the computational efficiency analysis. 
While GINE-KAN incurs significantly higher computation time per epoch compared to the base model (GINE), our model exhibits lower computation time.



\begin{table}[]
\centering
\begin{threeparttable}
\begin{tabular}{cccc}
\toprule[1pt]
\multicolumn{1}{c}{\begin{tabular}[c]{@{}c@{}}Agg.\\ with \\ SKAN\end{tabular}} & \multicolumn{1}{c}{\begin{tabular}[c]{@{}c@{}}Update\\ with \\ SKAN\end{tabular}} & \multicolumn{1}{c}{\begin{tabular}[c]{@{}c@{}}Classifier\\ with \\ SKAN\end{tabular}} & \multicolumn{1}{c}{\begin{tabular}[c]{@{}c@{}}Result\\ROC-AUC \end{tabular}}  \\
\midrule                                                         \text{\ding{55}} & \text{\ding{55}}  &  \text{\ding{55}} & 0.744 $\pm$ 0.014   
\\
\midrule
\text{\ding{51}} & \text{\ding{55}}  &  \text{\ding{55}}  &   0.746 $\pm$ 0.014  \\
\text{\ding{55}} & \text{\ding{51}}  &  \text{\ding{55}}  & \textcolor{springgreen}{0.764 $\pm$ 0.010}\tnote{a}    \\
\text{\ding{55}} & \text{\ding{55}}  &  \text{\ding{51}}  &     0.762 $\pm$ 0.008 \\
\midrule
\text{\ding{51}} & \text{\ding{51}}  &  \text{\ding{55}}  &    
0.767 $\pm$ 0.009 \\
\text{\ding{51}} & \text{\ding{55}}  &  \text{\ding{51}}  &   
0.761 $\pm$ 0.015\\
\text{\ding{55}} & \text{\ding{51}}  &  \text{\ding{51}}  & \textcolor{skyblue}{0.785 $\pm$ 0.009}     \\  
\midrule
\text{\ding{51}} & \text{\ding{51}}  &  \text{\ding{51}}  &   
\textcolor{orange}{0.782 $\pm$ 0.011}\tnote{b} \\
\bottomrule                                                      
\end{tabular}
\begin{tablenotes}[para, flushleft]
    \footnotesize
    \item[a]  GINE-SKAN.
    \item[b]  GINE-SKAN+.
\end{tablenotes}
\end{threeparttable}
\caption{Ablation study of SKAN in Aggregation and Update functions and classifier with GINE on the HIV dataset. The highlights are the \textcolor{skyblue}{first}, \textcolor{orange}{second}, and \textcolor{springgreen}{third} best results.}
\label{table:SKAN_position}
\end{table}

\paragraph{SKAN Integration in GNNs.}

Table~\ref{table:SKAN_position}  compares the performance of GINE with SKAN integrated into different components (Aggregation and Update, and Classifier) on the HIV dataset. The results are presented as the mean and standard deviation of the test ROC-AUC across five runs. The results indicate that integrating SKAN into the update function (GINE-SKAN) is more effective than integrating it into the aggregation function (GINE-SKAN (Agg.)). This advantage is likely due to the update function's role in executing nonlinear feature transformations at each GNN layer. Additionally, using SKAN as the classifier, our model (GINE-SKAN+) achieves state-of-the-art performance.

\section{Conclusion}

This work introduces GNN-SKAN and GNN-SKAN+, a new class of GNNs that integrate traditional GNN architectures with KAN. The core of our approach is SwallowKAN (SKAN), a novel KAN variant specifically designed to address the challenges of molecular diversity and enhance computational efficiency.
We evaluate our models on widely used molecular benchmark datasets, including six classification, six regression, and four few-shot learning datasets. The results show that our approach significantly enhances molecular representation learning.
In future work, we plan to explore the integration of Graph Transformer architectures with KAN to address the high computational complexity inherent in these models. This would potentially further enhance molecular representation learning tasks.

\nobibliography*

\bibliography{aaai25}

\begin{thebibliography}{64}
\providecommand{\natexlab}[1]{#1}

\bibitem[{Ahmad et~al.(2022)Ahmad, Simon, Chithrananda, Grand, and Ramsundar}]{ahmad2022chemberta}
Ahmad, W.; Simon, E.; Chithrananda, S.; Grand, G.; and Ramsundar, B. 2022.
\newblock Chemberta-2: Towards chemical foundation models.

\bibitem[{Arnold(2009)}]{arnold2009representation}
Arnold, V.~I. 2009.
\newblock On the representation of functions of several variables as a superposition of functions of a smaller number of variables.
\newblock \emph{Collected works: Representations of functions, celestial mechanics and KAM theory, 1957--1965}, 25--46.

\bibitem[{Azam and Akhtar(2024)}]{azam2024suitability}
Azam, B.; and Akhtar, N. 2024.
\newblock Suitability of KANs for Computer Vision: A preliminary investigation.

\bibitem[{Black et~al.(2023)Black, Wan, Nayyeri, and Wang}]{black2023understanding}
Black, M.; Wan, Z.; Nayyeri, A.; and Wang, Y. 2023.
\newblock Understanding oversquashing in gnns through the lens of effective resistance.
\newblock In \emph{International Conference on Machine Learning}, 2528--2547. PMLR.

\bibitem[{Braun and Griebel(2009)}]{braun2009constructive}
Braun, J.; and Griebel, M. 2009.
\newblock On a constructive proof of Kolmogorov’s superposition theorem.
\newblock \emph{Constructive approximation}, 30: 653--675.

\bibitem[{Bresson et~al.(2024)Bresson, Nikolentzos, Panagopoulos, Chatzianastasis, Pang, and Vazirgiannis}]{bresson2024kagnns}
Bresson, R.; Nikolentzos, G.; Panagopoulos, G.; Chatzianastasis, M.; Pang, J.; and Vazirgiannis, M. 2024.
\newblock Kagnns: Kolmogorov-arnold networks meet graph learning.

\bibitem[{Chen(2024)}]{chen2024gaussian}
Chen, A.~S. 2024.
\newblock Gaussian Process Kolmogorov-Arnold Networks.

\bibitem[{Chen et~al.(2024)Chen, Zhu, Zhang, Du, Li, Liu, Wu, and Wang}]{chen2024uncovering}
Chen, D.; Zhu, Y.; Zhang, J.; Du, Y.; Li, Z.; Liu, Q.; Wu, S.; and Wang, L. 2024.
\newblock Uncovering neural scaling laws in molecular representation learning.
\newblock \emph{Advances in Neural Information Processing Systems}, 36.

\bibitem[{Chithrananda, Grand, and Ramsundar(2020)}]{chithrananda2020chemberta}
Chithrananda, S.; Grand, G.; and Ramsundar, B. 2020.
\newblock ChemBERTa: large-scale self-supervised pretraining for molecular property prediction.

\bibitem[{Cybenko(1989)}]{cybenko1989approximation}
Cybenko, G. 1989.
\newblock Approximation by superpositions of a sigmoidal function.
\newblock \emph{Mathematics of control, signals and systems}, 2(4): 303--314.

\bibitem[{Datta and Grant(2004)}]{datta2004crystal}
Datta, S.; and Grant, D.~J. 2004.
\newblock Crystal structures of drugs: advances in determination, prediction and engineering.
\newblock \emph{Nature Reviews Drug Discovery}, 3(1): 42--57.

\bibitem[{De~Carlo, Mastropietro, and Anagnostopoulos(2024)}]{de2024kolmogorov}
De~Carlo, G.; Mastropietro, A.; and Anagnostopoulos, A. 2024.
\newblock Kolmogorov-arnold graph neural networks.

\bibitem[{Di~Giovanni et~al.(2023)Di~Giovanni, Giusti, Barbero, Luise, Lio, and Bronstein}]{di2023over}
Di~Giovanni, F.; Giusti, L.; Barbero, F.; Luise, G.; Lio, P.; and Bronstein, M.~M. 2023.
\newblock On over-squashing in message passing neural networks: The impact of width, depth, and topology.
\newblock In \emph{International Conference on Machine Learning}, 7865--7885. PMLR.

\bibitem[{Fabian et~al.(2020)Fabian, Edlich, Gaspar, Segler, Meyers, Fiscato, and Ahmed}]{fabian2020molecular}
Fabian, B.; Edlich, T.; Gaspar, H.; Segler, M.; Meyers, J.; Fiscato, M.; and Ahmed, M. 2020.
\newblock Molecular representation learning with language models and domain-relevant auxiliary tasks.

\bibitem[{Fang et~al.(2022)Fang, Liu, Lei, He, Zhang, Zhou, Wang, Wu, and Wang}]{fang2022geometry}
Fang, X.; Liu, L.; Lei, J.; He, D.; Zhang, S.; Zhou, J.; Wang, F.; Wu, H.; and Wang, H. 2022.
\newblock Geometry-enhanced molecular representation learning for property prediction.
\newblock \emph{Nature Machine Intelligence}, 4(2): 127--134.

\bibitem[{Fang et~al.(2024)Fang, Zhang, Chen, Guo, Fan, and Chen}]{fang2024domainagnostic}
Fang, Y.; Zhang, N.; Chen, Z.; Guo, L.; Fan, X.; and Chen, H. 2024.
\newblock Domain-Agnostic Molecular Generation with Chemical Feedback.
\newblock In \emph{The Twelfth International Conference on Learning Representations}.

\bibitem[{Fang et~al.(2023)Fang, Zhang, Zhang, Chen, Zhuang, Shao, Fan, and Chen}]{fang2023knowledge}
Fang, Y.; Zhang, Q.; Zhang, N.; Chen, Z.; Zhuang, X.; Shao, X.; Fan, X.; and Chen, H. 2023.
\newblock Knowledge graph-enhanced molecular contrastive learning with functional prompt.
\newblock \emph{Nature Machine Intelligence}, 5(5): 542--553.

\bibitem[{Finn, Abbeel, and Levine(2017)}]{finn2017model}
Finn, C.; Abbeel, P.; and Levine, S. 2017.
\newblock Model-agnostic meta-learning for fast adaptation of deep networks.
\newblock In \emph{International conference on machine learning}, 1126--1135. PMLR.

\bibitem[{Gilmer et~al.(2020)Gilmer, Schoenholz, Riley, Vinyals, and Dahl}]{gilmer2020message}
Gilmer, J.; Schoenholz, S.~S.; Riley, P.~F.; Vinyals, O.; and Dahl, G.~E. 2020.
\newblock Message passing neural networks.
\newblock \emph{Machine learning meets quantum physics}, 199--214.

\bibitem[{Guo et~al.(2022)Guo, Shou, Makatura, Erps, Foshey, and Matusik}]{guo2022polygrammar}
Guo, M.; Shou, W.; Makatura, L.; Erps, T.; Foshey, M.; and Matusik, W. 2022.
\newblock Polygrammar: grammar for digital polymer representation and generation.
\newblock \emph{Advanced Science}, 9(23): 2101864.

\bibitem[{Hou et~al.(2022)Hou, Liu, Cen, Dong, Yang, Wang, and Tang}]{hou2022graphmae}
Hou, Z.; Liu, X.; Cen, Y.; Dong, Y.; Yang, H.; Wang, C.; and Tang, J. 2022.
\newblock Graphmae: Self-supervised masked graph autoencoders.
\newblock In \emph{Proceedings of the 28th ACM SIGKDD Conference on Knowledge Discovery and Data Mining}, 594--604.

\bibitem[{Hu* et~al.(2020)Hu*, Liu*, Gomes, Zitnik, Liang, Pande, and Leskovec}]{Hu*2020Strategies}
Hu*, W.; Liu*, B.; Gomes, J.; Zitnik, M.; Liang, P.; Pande, V.; and Leskovec, J. 2020.
\newblock Strategies for Pre-training Graph Neural Networks.
\newblock In \emph{International Conference on Learning Representations}.

\bibitem[{Jablonka et~al.(2024)Jablonka, Schwaller, Ortega-Guerrero, and Smit}]{jablonka2024leveraging}
Jablonka, K.~M.; Schwaller, P.; Ortega-Guerrero, A.; and Smit, B. 2024.
\newblock Leveraging large language models for predictive chemistry.
\newblock \emph{Nature Machine Intelligence}, 6(2): 161--169.

\bibitem[{Keles, Wijewardena, and Hegde(2023)}]{keles2023computational}
Keles, F.~D.; Wijewardena, P.~M.; and Hegde, C. 2023.
\newblock On the computational complexity of self-attention.
\newblock In \emph{International Conference on Algorithmic Learning Theory}, 597--619. PMLR.

\bibitem[{Kiamari, Kiamari, and Krishnamachari(2024)}]{kiamari2024gkan}
Kiamari, M.; Kiamari, M.; and Krishnamachari, B. 2024.
\newblock GKAN: Graph Kolmogorov-Arnold Networks.

\bibitem[{Kipf and Welling(2017)}]{kipf2017semisupervised}
Kipf, T.~N.; and Welling, M. 2017.
\newblock Semi-Supervised Classification with Graph Convolutional Networks.
\newblock In \emph{International Conference on Learning Representations}.

\bibitem[{Kolmogorov(1957)}]{kolmogorov1957presentation}
Kolmogorov, A.~N. 1957.
\newblock On the representation of continuous functions of several variables as superpositions of continuous functions of one variable and addition.
\newblock In \emph{Reports of the Academy of Sciences}, volume 114, 953--956. Russian Academy of Sciences.

\bibitem[{Li et~al.(2024)Li, Liu, Li, Wang, Liu, and Yuan}]{li2024u}
Li, C.; Liu, X.; Li, W.; Wang, C.; Liu, H.; and Yuan, Y. 2024.
\newblock U-KAN Makes Strong Backbone for Medical Image Segmentation and Generation.

\bibitem[{Li(2024)}]{li2024kolmogorov}
Li, Z. 2024.
\newblock Kolmogorov-arnold networks are radial basis function networks.

\bibitem[{Li et~al.(2022)Li, Jiang, Wang, and Zhang}]{li2022deep}
Li, Z.; Jiang, M.; Wang, S.; and Zhang, S. 2022.
\newblock Deep learning methods for molecular representation and property prediction.
\newblock \emph{Drug Discovery Today}, 27(12): 103373.

\bibitem[{Liu et~al.(2018)Liu, Lee, Park, Kim, Yang, Hwang, and Yang}]{liu2018learning}
Liu, Y.; Lee, J.; Park, M.; Kim, S.; Yang, E.; Hwang, S.~J.; and Yang, Y. 2018.
\newblock Learning to propagate labels: Transductive propagation network for few-shot learning.
\newblock \emph{arXiv preprint arXiv:1805.10002}.

\bibitem[{Liu et~al.(2024)Liu, Wang, Vaidya, Ruehle, Halverson, Solja{\v{c}}i{\'c}, Hou, and Tegmark}]{liu2024kan}
Liu, Z.; Wang, Y.; Vaidya, S.; Ruehle, F.; Halverson, J.; Solja{\v{c}}i{\'c}, M.; Hou, T.~Y.; and Tegmark, M. 2024.
\newblock Kan: Kolmogorov-arnold networks.
\newblock \emph{arXiv preprint arXiv:2404.19756}.

\bibitem[{Livingstone(2000)}]{livingstone2000characterization}
Livingstone, D.~J. 2000.
\newblock The characterization of chemical structures using molecular properties. A survey.
\newblock \emph{Journal of chemical information and computer sciences}, 40(2): 195--209.

\bibitem[{Mayr et~al.(2018)Mayr, Klambauer, Unterthiner, Steijaert, Wegner, Ceulemans, Clevert, and Hochreiter}]{mayr2018large}
Mayr, A.; Klambauer, G.; Unterthiner, T.; Steijaert, M.; Wegner, J.~K.; Ceulemans, H.; Clevert, D.-A.; and Hochreiter, S. 2018.
\newblock Large-scale comparison of machine learning methods for drug target prediction on ChEMBL.
\newblock \emph{Chemical science}, 9(24): 5441--5451.

\bibitem[{Pei et~al.(2024)Pei, Chen, Chen, Deng, Tao, Wang, and Guan}]{pei2024hago}
Pei, H.; Chen, T.; Chen, A.; Deng, H.; Tao, J.; Wang, P.; and Guan, X. 2024.
\newblock Hago-net: Hierarchical geometric massage passing for molecular representation learning.
\newblock In \emph{Proceedings of the AAAI Conference on Artificial Intelligence}, volume~38, 14572--14580.

\bibitem[{Pilania(2021)}]{pilania2021machine}
Pilania, G. 2021.
\newblock Machine learning in materials science: From explainable predictions to autonomous design.
\newblock \emph{Computational Materials Science}, 193: 110360.

\bibitem[{Ramp{\'a}{\v{s}}ek et~al.(2022)Ramp{\'a}{\v{s}}ek, Galkin, Dwivedi, Luu, Wolf, and Beaini}]{rampavsek2022recipe}
Ramp{\'a}{\v{s}}ek, L.; Galkin, M.; Dwivedi, V.~P.; Luu, A.~T.; Wolf, G.; and Beaini, D. 2022.
\newblock Recipe for a general, powerful, scalable graph transformer.
\newblock \emph{Advances in Neural Information Processing Systems}, 35: 14501--14515.

\bibitem[{Rong et~al.(2020)Rong, Bian, Xu, Xie, Wei, Huang, and Huang}]{rong2020self}
Rong, Y.; Bian, Y.; Xu, T.; Xie, W.; Wei, Y.; Huang, W.; and Huang, J. 2020.
\newblock Self-supervised graph transformer on large-scale molecular data.
\newblock \emph{Advances in neural information processing systems}, 33: 12559--12571.

\bibitem[{Sadybekov and Katritch(2023)}]{sadybekov2023computational}
Sadybekov, A.~V.; and Katritch, V. 2023.
\newblock Computational approaches streamlining drug discovery.
\newblock \emph{Nature}, 616(7958): 673--685.

\bibitem[{Seeger(2004)}]{seeger2004gaussian}
Seeger, M. 2004.
\newblock Gaussian processes for machine learning.
\newblock \emph{International journal of neural systems}, 14(02): 69--106.

\bibitem[{Snell, Swersky, and Zemel(2017)}]{snell2017prototypical}
Snell, J.; Swersky, K.; and Zemel, R. 2017.
\newblock Prototypical networks for few-shot learning.
\newblock \emph{Advances in neural information processing systems}, 30.

\bibitem[{Song et~al.(2020)Song, Zheng, Niu, Fu, Lu, and Yang}]{song2020communicative}
Song, Y.; Zheng, S.; Niu, Z.; Fu, Z.-H.; Lu, Y.; and Yang, Y. 2020.
\newblock Communicative Representation Learning on Attributed Molecular Graphs.
\newblock In \emph{IJCAI}, volume 2020, 2831--2838.

\bibitem[{Ta(2024)}]{ta2024bsrbf}
Ta, H.-T. 2024.
\newblock BSRBF-KAN: A combination of B-splines and Radial Basic Functions in Kolmogorov-Arnold Networks.

\bibitem[{Unser, Aldroubi, and Eden(1993)}]{unser1993b}
Unser, M.; Aldroubi, A.; and Eden, M. 1993.
\newblock B-spline signal processing. I. Theory.
\newblock \emph{IEEE transactions on signal processing}, 41(2): 821--833.

\bibitem[{Vaca-Rubio et~al.(2024)Vaca-Rubio, Blanco, Pereira, and Caus}]{vaca2024kolmogorov}
Vaca-Rubio, C.~J.; Blanco, L.; Pereira, R.; and Caus, M. 2024.
\newblock Kolmogorov-arnold networks (kans) for time series analysis.

\bibitem[{Vaswani et~al.(2017)Vaswani, Shazeer, Parmar, Uszkoreit, Jones, Gomez, Kaiser, and Polosukhin}]{vaswani2017attention}
Vaswani, A.; Shazeer, N.; Parmar, N.; Uszkoreit, J.; Jones, L.; Gomez, A.~N.; Kaiser, {\L}.; and Polosukhin, I. 2017.
\newblock Attention is all you need.
\newblock \emph{Advances in neural information processing systems}, 30.

\bibitem[{Veli{\v{c}}kovi{\'c} et~al.(2018)Veli{\v{c}}kovi{\'c}, Cucurull, Casanova, Romero, Li{\`o}, and Bengio}]{velivckovic2018graph}
Veli{\v{c}}kovi{\'c}, P.; Cucurull, G.; Casanova, A.; Romero, A.; Li{\`o}, P.; and Bengio, Y. 2018.
\newblock Graph Attention Networks.
\newblock In \emph{International Conference on Learning Representations}.

\bibitem[{Wang et~al.(2021{\natexlab{a}})Wang, ABUDUWEILI, quanming yao, and Dou}]{wang2021propertyaware}
Wang, Y.; ABUDUWEILI, A.; quanming yao; and Dou, D. 2021{\natexlab{a}}.
\newblock Property-Aware Relation Networks for Few-Shot Molecular Property Prediction.
\newblock In Beygelzimer, A.; Dauphin, Y.; Liang, P.; and Vaughan, J.~W., eds., \emph{Advances in Neural Information Processing Systems}.

\bibitem[{Wang et~al.(2021{\natexlab{b}})Wang, Abuduweili, Yao, and Dou}]{wang2021property}
Wang, Y.; Abuduweili, A.; Yao, Q.; and Dou, D. 2021{\natexlab{b}}.
\newblock Property-aware relation networks for few-shot molecular property prediction.
\newblock \emph{Advances in Neural Information Processing Systems}, 34: 17441--17454.

\bibitem[{Wang et~al.(2022)Wang, Wang, Cao, and Barati~Farimani}]{wang2022molecular}
Wang, Y.; Wang, J.; Cao, Z.; and Barati~Farimani, A. 2022.
\newblock Molecular contrastive learning of representations via graph neural networks.
\newblock \emph{Nature Machine Intelligence}, 4(3): 279--287.

\bibitem[{Wang et~al.(2024)Wang, Yu, Gao, Sha, Wang, Gao, Zhang, and Rong}]{wang2024spectralkan}
Wang, Y.; Yu, X.; Gao, Y.; Sha, J.; Wang, J.; Gao, L.; Zhang, Y.; and Rong, X. 2024.
\newblock SpectralKAN: Kolmogorov-Arnold Network for Hyperspectral Images Change Detection.

\bibitem[{Wieder et~al.(2020)Wieder, Kohlbacher, Kuenemann, Garon, Ducrot, Seidel, and Langer}]{wieder2020compact}
Wieder, O.; Kohlbacher, S.; Kuenemann, M.; Garon, A.; Ducrot, P.; Seidel, T.; and Langer, T. 2020.
\newblock A compact review of molecular property prediction with graph neural networks.
\newblock \emph{Drug Discovery Today: Technologies}, 37: 1--12.

\bibitem[{Wu et~al.(2012)Wu, Wang, Zhang, and Du}]{wu2012using}
Wu, Y.; Wang, H.; Zhang, B.; and Du, K.-L. 2012.
\newblock Using radial basis function networks for function approximation and classification.
\newblock \emph{International Scholarly Research Notices}, 2012(1): 324194.

\bibitem[{Xiong et~al.(2019)Xiong, Wang, Liu, Zhong, Wan, Li, Li, Luo, Chen, Jiang et~al.}]{xiong2019pushing}
Xiong, Z.; Wang, D.; Liu, X.; Zhong, F.; Wan, X.; Li, X.; Li, Z.; Luo, X.; Chen, K.; Jiang, H.; et~al. 2019.
\newblock Pushing the boundaries of molecular representation for drug discovery with the graph attention mechanism.
\newblock \emph{Journal of medicinal chemistry}, 63(16): 8749--8760.

\bibitem[{Xu, Chen, and Wang(2024)}]{xu2024kolmogorov}
Xu, K.; Chen, L.; and Wang, S. 2024.
\newblock Kolmogorov-Arnold Networks for Time Series: Bridging Predictive Power and Interpretability.

\bibitem[{Xu et~al.(2018)Xu, Hu, Leskovec, and Jegelka}]{xu2018powerful}
Xu, K.; Hu, W.; Leskovec, J.; and Jegelka, S. 2018.
\newblock How Powerful are Graph Neural Networks?
\newblock In \emph{International Conference on Learning Representations}.

\bibitem[{Yang et~al.(2019)Yang, Swanson, Jin, Coley, Eiden, Gao, Guzman-Perez, Hopper, Kelley, Mathea et~al.}]{yang2019analyzing}
Yang, K.; Swanson, K.; Jin, W.; Coley, C.; Eiden, P.; Gao, H.; Guzman-Perez, A.; Hopper, T.; Kelley, B.; Mathea, M.; et~al. 2019.
\newblock Analyzing learned molecular representations for property prediction.

\bibitem[{Yi et~al.(2022)Yi, You, Huang, and Kwoh}]{yi2022graph}
Yi, H.-C.; You, Z.-H.; Huang, D.-S.; and Kwoh, C.~K. 2022.
\newblock Graph representation learning in bioinformatics: trends, methods and applications.
\newblock \emph{Briefings in Bioinformatics}, 23(1): bbab340.

\bibitem[{Ying et~al.(2021)Ying, Cai, Luo, Zheng, Ke, He, Shen, and Liu}]{ying2021transformers}
Ying, C.; Cai, T.; Luo, S.; Zheng, S.; Ke, G.; He, D.; Shen, Y.; and Liu, T.-Y. 2021.
\newblock Do transformers really perform badly for graph representation?
\newblock \emph{Advances in neural information processing systems}, 34: 28877--28888.

\bibitem[{Yu and Gao(2022)}]{yu2022molecular}
Yu, Z.; and Gao, H. 2022.
\newblock Molecular representation learning via heterogeneous motif graph neural networks.
\newblock In \emph{International Conference on Machine Learning}, 25581--25594. PMLR.

\bibitem[{Y{\"u}ksel et~al.(2023)Y{\"u}ksel, Ulusoy, {\"U}nl{\"u}, and Do{\u{g}}an}]{yuksel2023selformer}
Y{\"u}ksel, A.; Ulusoy, E.; {\"U}nl{\"u}, A.; and Do{\u{g}}an, T. 2023.
\newblock SELFormer: molecular representation learning via SELFIES language models.
\newblock \emph{Machine Learning: Science and Technology}, 4(2): 025035.

\bibitem[{Zhang and Zhang(2024)}]{zhang2024graphkan}
Zhang, F.; and Zhang, X. 2024.
\newblock GraphKAN: Enhancing Feature Extraction with Graph Kolmogorov Arnold Networks.

\bibitem[{Zhang et~al.(2021)Zhang, Liu, Wang, Lu, and Lee}]{zhang2021motif}
Zhang, Z.; Liu, Q.; Wang, H.; Lu, C.; and Lee, C.-K. 2021.
\newblock Motif-based graph self-supervised learning for molecular property prediction.
\newblock \emph{Advances in Neural Information Processing Systems}, 34: 15870--15882.

\bibitem[{Zhou et~al.(2023)Zhou, Gao, Ding, Zheng, Xu, Wei, Zhang, and Ke}]{zhou2023unimol}
Zhou, G.; Gao, Z.; Ding, Q.; Zheng, H.; Xu, H.; Wei, Z.; Zhang, L.; and Ke, G. 2023.
\newblock Uni-Mol: A Universal 3D Molecular Representation Learning Framework.
\newblock In \emph{The Eleventh International Conference on Learning Representations}.

\end{thebibliography}

\appendix

\end{document}


\maketitle

\linenumbers

\appendix 
\onecolumn
\setcounter{table}{0}
\setcounter{figure}{0}
\renewcommand{\thetable}{A\arabic{table}}  
\renewcommand{\thefigure}{A\arabic{figure}}
\setcounter{equation}{0}
\renewcommand{\theequation}{A\arabic{equation}}

\section{Benchmark Datasets}

\begin{table*}[htb]
    \centering
    \resizebox{\linewidth}{!}{
    \begin{tabular}{lllll}
    \toprule[1pt]
    \multicolumn{1}{l}{\textbf{Dataset}} & \multicolumn{1}{l}{\textbf{Type of the task}} & \multicolumn{1}{l}{\textbf{Content}} & \multicolumn{1}{l}{\textbf{\# of compounds}} & \multicolumn{1}{l}{\textbf{Metric}}  \\
\midrule 
BBBP  &  Binary classification  & Binary labels on blood-brain barrier permeability & 2,039 & ROC-AUC  \\
Tox21 &  Binary classification & Qualitative toxicity measurements on 12 targets & 7,831 & ROC-AUC \\
ToxCast &  Binary classification & Qualitative toxicity measurements on 617 targets  & 8,597  & ROC-AUC \\
SIDER &  Binary classification & Classification of drug side-effects into 27 system organ classes  & 1,427 & ROC-AUC  \\ 
HIV & Binary classification  & Binary labels on the ability to inhibit HIV replication & 41,127 & ROC-AUC   \\
BACE & Binary classification & Binary labels on human-secretase 1 (BACE1) binding properties & 1,513 & ROC-AUC \\
\midrule
Lipo &  Regression & Experimental octanol/water distribution coefficient of compounds & 4,200 & MAE  \\
FreeSolv & Regression & Hydration free energy of small molecules in water & 642 & MAE  \\ 
Esol &  Regression & Aqueous solubility of common small molecules & 1,128 & MAE \\ 
QM7 & Regression & Prediction of atomization energies & 7,165 & MAE \\ 
QM8 & Regression  & Prediction of 12 quantum mechanical properties & 21,786 & MAE  \\
QM9 & Regression & Prediction of 12 quantum mechanical properties & 134,000 & MAE \\
\bottomrule
\end{tabular}}
    \caption{Detailed description of molecular classification and regression datasets.}
    \label{tab:benchmark_and_dataset}
\end{table*}

\begin{table*}[htb]
\centering
\begin{tabular}{llccc}
\toprule[1pt]
\multicolumn{1}{l}{\textbf{Dataset}} & \multicolumn{1}{l}{\textbf{\# Compounds}} & \multicolumn{1}{c}{\textbf{\# Tasks}} & \multicolumn{1}{c}{\textbf{\# Meta-Training Tasks}} & \multicolumn{1}{c}{\textbf{\# Meta-Testing Tasks}}  \\
\midrule 
Tox21 & 8,014 & 12 & 9 & 3  \\
SIDER & 1,427 & 27 & 21 & 6  \\ 
MUV & 93,127 & 17 & 12 & 5 \\ 
ToxCast & 8,597 & 617 & 450 & 167 \\ 
\bottomrule
\end{tabular}
\caption{Summary of few-shot datasets used.}
\label{tab: few_shot_datasets}
\end{table*}
Table \ref{tab:benchmark_and_dataset} provides a detailed breakdown of various molecular datasets used for molecular classification and regression tasks. It lists the dataset name, the type of task (binary classification or regression), the content (a brief description of the specific task or property being predicted), the number of compounds included in each dataset, and the evaluation metric used (ROC-AUC for classification tasks and MAE for regression tasks). 

Table \ref{tab: few_shot_datasets} summarizes datasets that are employed in few-shot learning experiments \cite{wang2021propertyaware}. It details the number of compounds, the total number of tasks, and further breaks down the tasks into meta-training and meta-testing tasks for each dataset. The few-shot datasets include Tox21, SIDER, MUV, and ToxCast, with varying numbers of compounds and tasks.

\section{Comparison with SOTAs} 
\begin{table*}[htb]
\centering
\begin{tabular}{lllllll}
\toprule[1pt]
\multicolumn{1}{l}{\textbf{Model}} & \multicolumn{1}{c}{\textbf{BBBP} $\uparrow$}       & \multicolumn{1}{c}{\textbf{Tox21} $\uparrow$}      & \multicolumn{1}{c}{\textbf{ToxCast} $\uparrow$}    & \multicolumn{1}{c}{\textbf{SIDER} $\uparrow$}      & \multicolumn{1}{c}{\textbf{HIV} $\uparrow$}        & \multicolumn{1}{c}{\textbf{BACE} $\uparrow$}       \\ 
\midrule
GCN                       & 0.622 $\pm$ 0.037 & 0.738 $\pm$ 0.003 & 0.624 $\pm$ 0.006 & 0.611 $\pm$ 0.012 & 0.748 $\pm$ 0.007 & 0.696 $\pm$ 0.048 \\
GAT                       & 0.640 $\pm$ 0.032 & 0.732 $\pm$ 0.003 & 0.635 $\pm$ 0.012 & 0.578 $\pm$ 0.017 & 0.751 $\pm$ 0.010 & 0.653 $\pm$ 0.016 \\
GINE                    & 0.644 $\pm$ 0.018 & 0.733 $\pm$ 0.009 & 0.612 $\pm$ 0.009 & 0.596 $\pm$ 0.003 & 0.744 $\pm$ 0.014 & 0.612 $\pm$ 0.056 \\
CMPNN  & \textbf{0.737 $\pm$ 0.065}     &   0.709 $\pm$  0.024                
&   \textbf{0.708 $\pm$ 0.013}                   &   0.548 $\pm$ 0.020                   &   0.782 $\pm$ 0.022                   &   0.784 $\pm$ 0.026                             \\ 
\midrule
MolCLR 
&  0.706 $\pm$ 0.031    
&  0.747 $\pm$ 0.080                     
&  0.659 $\pm$ 0.021                    
&  0.612 $\pm$ 0.036                     
& 0.745 $\pm$ 0.210            & 0.823 $\pm$ 0.012                               \\
MGSSL  &  0.724 $\pm$ 0.070         &  0.745 $\pm$ 0.026           &  0.638 $\pm$ 0.023           & 0.578 $\pm$ 0.036            &  0.774 $\pm$ 0.006           & 0.824 $\pm$ 0.021                  \\
GraphMAE  & 0.706 $\pm$ 0.060   & 0.744 $\pm$ 0.034                           & 0.641 $\pm$ 0.003          & 0.603 $\pm$ 0.011          & 0.772 $\pm$ 0.010          & 0.780 $\pm$ 0.012 
\\ 
\midrule
Grover$_{base}$
& 0.700 $\pm$ 0.009 
& 0.743 $\pm$ 0.012 
& 0.654 $\pm$ 0.006 
& \textbf{0.656 $\pm$ 0.005} 
& 0.776 $\pm$ 0.023 
& \textbf{0.826 $\pm$ 0.016} \\ 
GPS 
& 0.698 $\pm$ 0.012 
& 0.748 $\pm$ 0.006 
& 0.655 $\pm$ 0.014 
& 0.645 $\pm$ 0.008 
& 0.774 $\pm$ 0.005 
& 0.819 $\pm$ 0.011 \\
\midrule
GNN-SKAN                    & 0.652 $\pm$ 0.009 & 0.738 $\pm$ 0.008 & 0.633 $\pm$ 0.003 & 0.603 $\pm$ 0.004 & 0.764 $\pm$ 0.008 & 0.768 $\pm$ 0.011 \\
GNN-SKAN+                   & 0.688 $\pm$ 0.015 & \textbf{0.750 $\pm$ 0.002} & 0.643 $\pm$ 0.004 & 0.620 $\pm$ 0.002 & \textbf{0.786 $\pm$ 0.015} & 0.762 $\pm$ 0.027 \\
\bottomrule[1pt]
\end{tabular}
\caption{Comparison of our best results from Table 1 with the SOTA models on six classification benchmarks of physiology and biophysics. The best results are indicated in bold. The mean and standard deviation of test ROC-AUC on five independent runs are reported.}
\label{table:classification_sota}
\end{table*}

Table~\ref{table:classification_sota} compares the best results of our augmented models from Table 1 with the SOTAs on six molecular classification datasets.  The baselines are divided into three categories: 1) supervised methods: GCN \cite{kipf2017semisupervised}, GAT \cite{velivckovic2018graph}, GINE \cite{Hu*2020Strategies}, and CMPNN \cite{song2020communicative}; 2) Self-supervised methods: MolCLR \cite{wang2022molecular}, MGSSL \cite{zhang2021motif}, and GraphMAE \cite{hou2022graphmae}; 3) Graph Transformers: Grover$_{base}$ \cite{rong2020self} and GPS \cite{rampavsek2022recipe}. The results demonstrate that our augmented models match or surpass the SOTAs. 

\begin{figure}
    \centering
    \includegraphics[width=0.8\linewidth]{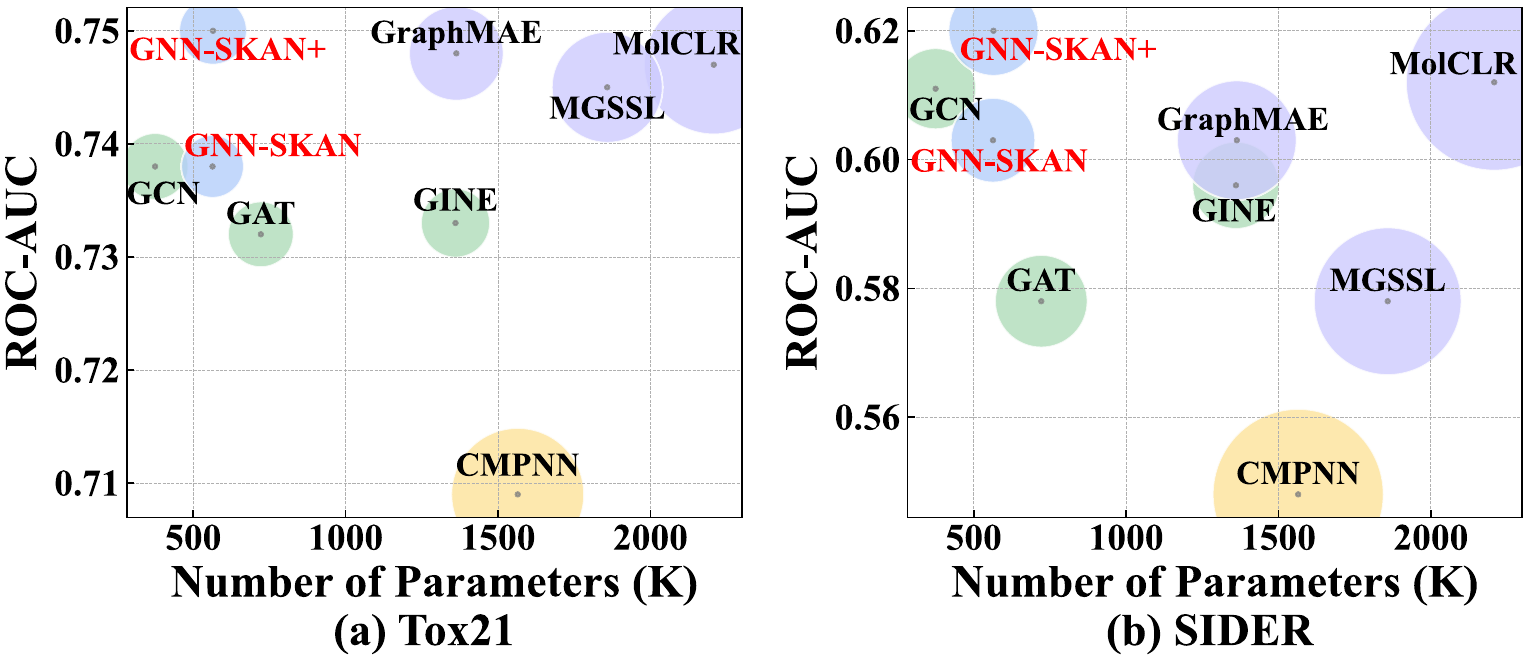}
    \caption{Performance comparison of our models with the SOTAs on the Tox21 and SIDER datasets. The x-axis represents the number of parameters (K), and the y-axis represents the ROC-AUC scores. Each circle's size corresponds to the per-epoch time. GNN-SKAN+ and GNN-SKAN are highlighted in red. Purple circles represent SSL methods, green and yellow circles represent supervised learning methods, and blue circles represent our methods.}
    \label{fig:enter-label}
\end{figure}

Figure~\ref{fig:enter-label} illustrates that GNN-SKAN+ is a strong contender in molecular classification tasks, offering a compelling balance of high performance, low parameter count, and computational efficiency. This makes it a versatile choice for applications where both accuracy and resource efficiency are critical. In contrast, while models like MolCLR \cite{wang2022molecular} and GraphMAE \cite{hou2022graphmae} achieve high ROC-AUC scores, their increased complexity and computational costs may limit their practicality in some scenarios.

\section{Ablation Study}

\subsection{Architectural Layers}
\begin{figure}[htb]
    \centering
\includegraphics[width=0.5\linewidth]{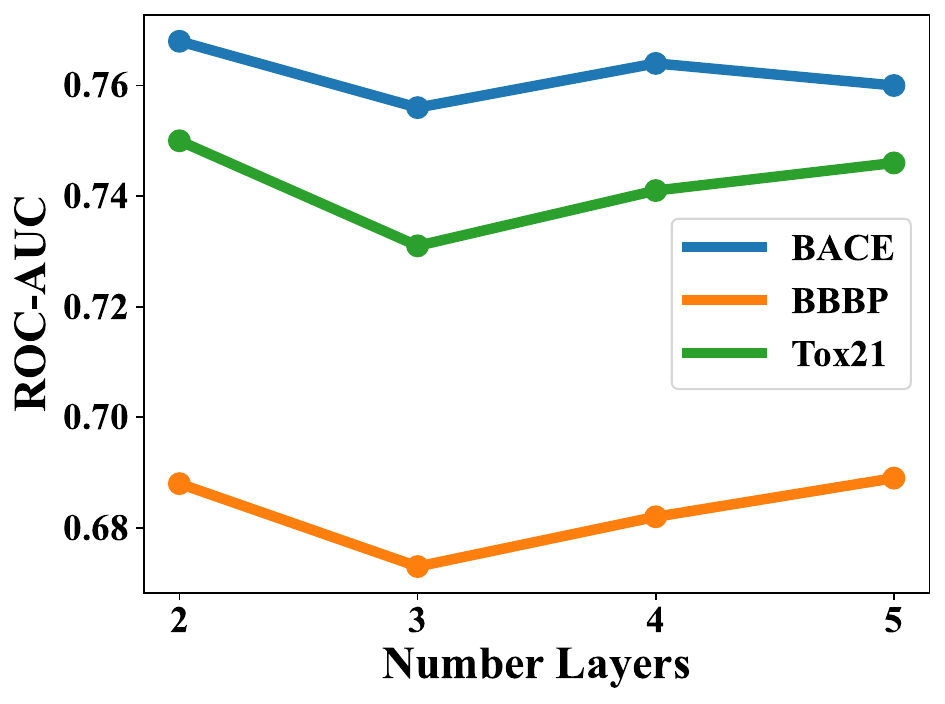}
    \caption{The performance of increasing number of  GINE-SKAN layers.}
    \label{fig:ablation_gine_scan}
\end{figure}

Figure~\ref{fig:ablation_gine_scan} illustrates the impact of increasing the number of GINE-SKAN layers on model performance. On the BACE dataset, our model consistently showing higher ROC-AUC values across all layer configurations. On the Tox21 dataset, our model exhibits stable performance, with a slight dip at 3 layers. Meanwhile, on the BBBP dataset, our model shows an increase in performance as the number of layers increases. Based on these observations, we ultimately set the number of GINE-SKAN layers to 2. 

\begin{figure}
    \centering
    \includegraphics[width=0.5\linewidth]{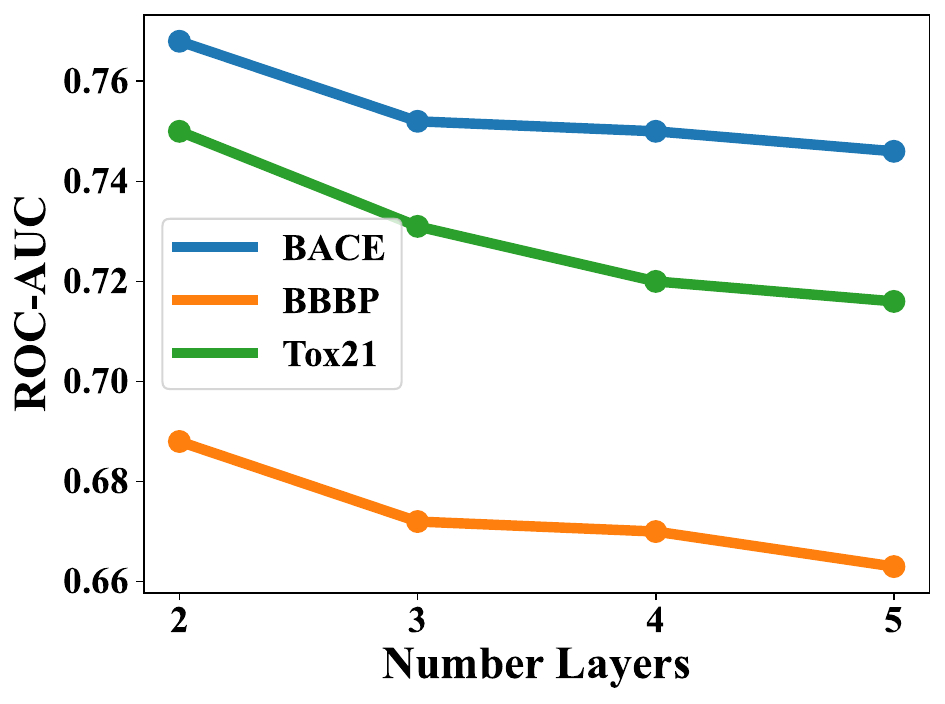}
    \caption{The performance of increasing SKAN layers.}
    \label{fig:scan_layer}
\end{figure}

Figure \ref{fig:scan_layer} demonstrates that increasing the number of SKAN layers in the GINE-SKAN model does not necessarily lead to better performance. In fact, for all three datasets, there is a general trend of declining ROC-AUC scores as the model becomes deeper. This suggests that while a modest number of layers (around 2-3) may be beneficial, further increasing the depth of the model could introduce overfitting or unnecessary complexity, ultimately reducing the model's effectiveness. Therefore, careful consideration of the number of layers is crucial when designing GINE-SKAN architectures for different datasets. Based on these observations, we ultimately set the number of SKAN layers to 2.

\subsection{Hyperparameter Settings}

\begin{table*}
\centering
\begin{tabular}{lc}
\toprule[1pt]
\multicolumn{1}{l}{Hyperparameter} & \multicolumn{1}{c}{Selected} \\
\midrule
learning rate & [0.001, 0.0001] \\
number of GNN-SKAN layers & 2 \\
number of SKAN layers & 2 \\
number of Adaptive RBFs & [3,4,5,6,7,8] \\
bandwidth & [2, 4, 6, 8] \\ 
dropout & 0 \\ 
hidden dimension for GNN-SKAN & 256 \\
optimizer & Adam \\
\bottomrule[1pt]
\end{tabular}
\caption{Hyperparameters used by GNN-SKAN.}
\end{table*}

The hyperparameters selected for the GNN-SKAN model reflect a deliberate balance between simplicity and performance. With 2 layers each for GNN-SKAN and SKAN, no dropout, and a hidden dimension of 256, the model is designed to avoid overfitting while still being capable of capturing complex patterns. The use of a low learning rate and the Adam optimizer further supports the goal of achieving stable and effective training. This setup suggests that the GNN-SKAN model is optimized for scenarios where model simplicity and computational efficiency are important, without sacrificing too much in terms of performance.

\nobibliography*

\bibliography{aaai25}